\documentclass[lettersize,journal]{IEEEtran}
\usepackage{amsmath,amsfonts}
\usepackage{algorithmic}
\usepackage{array}
\usepackage[caption=false,font=normalsize,labelfont=sf,textfont=sf]{subfig}
\usepackage{textcomp}
\usepackage{stfloats}
\usepackage{url}
\usepackage{verbatim}
\usepackage{graphicx}
\usepackage{amssymb}
\usepackage{mathtools}
\usepackage{bm}
\usepackage{booktabs}
\usepackage{multirow}
\usepackage{xcolor}
\usepackage{colortbl}
\usepackage{microtype}
\usepackage{siunitx}
\usepackage{amsthm}
\begin{document}

\title{StreamPPG: Low-Latency rPPG Estimation \\ via Consistent Privileged Learning}

\author{Yiming Li, Yihan Yang, Yuguang Chu, Yuanhui Hu, Lu Yu, Xiaohan Zhang, and Hui-Liang Shen,~\IEEEmembership{Senior Member,~IEEE}

}

\markboth{arxiv}%
{Shell \MakeLowercase{\textit{et al.}}: A Sample Article Using IEEEtran.cls for IEEE Journals}


\maketitle
\begin{abstract}
Remote photoplethysmography (rPPG) estimates the blood volume pulse (BVP) signal from facial videos, enabling contact-free health monitoring. Conventional clip-wise approaches, which use video clips as input, require capturing over one hundred frames before inference, thus introducing several seconds of delay and hindering real-time use. Meanwhile, frame-wise approaches struggle to capture long-range temporal and periodic features of physiological rhythms, and therefore lead to reduced estimation accuracy. To overcome these issues, we propose StreamPPG, a unified architecture that enables low-latency frame-wise physiological signal estimation while achieving competitive accuracy compared with clip-wise approaches. StreamPPG is trained under a consistent privileged learning (CPL) strategy, which leverages ground-truth rPPG signals as privileged information to enhance the model's representation capability. Extensive experiments demonstrate that StreamPPG achieves state-of-the-art accuracy across multiple datasets while maintaining real-time throughput on edge devices.
\end{abstract}


\section{Introduction}
\label{sec:intro}
Photoplethysmography (PPG) is a physiological signal commonly used to estimate vital parameters such as heart rate, respiration, and blood oxygen saturation. However, conventional PPG relies on direct skin contact, which poses infection risks in scenarios such as postoperative monitoring or treatment of patients with skin injuries and burns \cite{sun2015photoplethysmography}. To address this issue, remote photoplethysmography (rPPG) has been developed to estimate physiological signals from facial videos by analyzing subtle color fluctuations induced by blood volume changes \cite{Green}. This technique enables safe and convenient physiological monitoring, greatly expanding the applicability of PPG to remote healthcare \cite{yan2018contact, jiang2025lsts}, affective computing \cite{yu2021facial}, driver monitoring \cite{savic2025rs+}. However, the accuracy and real-time performance of rPPG signal prediction remain significant challenges.

\begin{figure}[th]
    \centering
    \includegraphics[width=0.9\columnwidth]{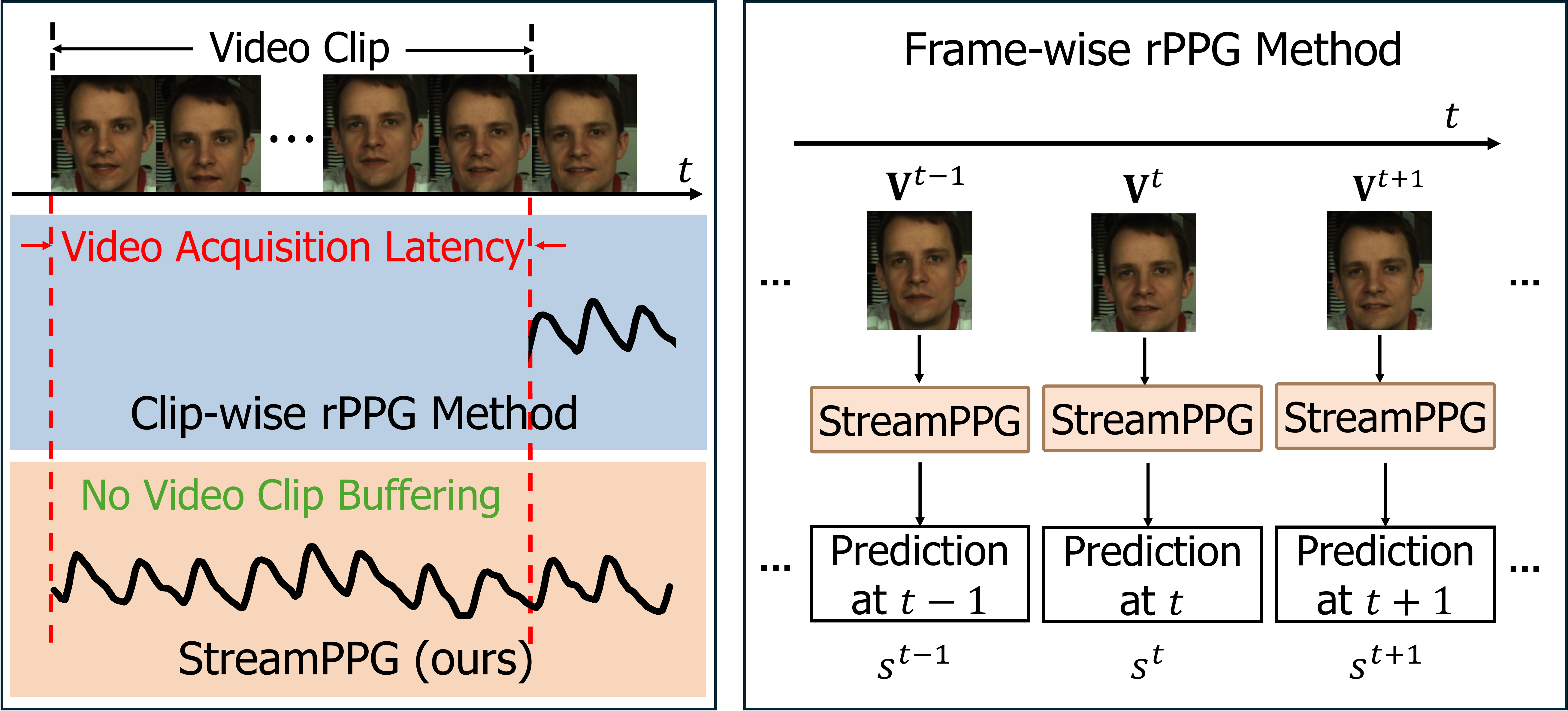}
    \caption{Comparison between clip-wise rPPG inference and the proposed StreamPPG frame-wise streaming inference. Unlike clip-wise rPPG methods that require buffering a complete video clip before inference, StreamPPG processes incoming frames sequentially and predicts the corresponding rPPG waveform at each time step, enabling causal frame-wise streaming estimation without video clip buffering.}
    \label{fig:computational cost}
\end{figure}

Accurate measurement of rPPG signals remains challenging because the subtle skin color variations caused by blood volume changes are easily overwhelmed by lighting fluctuations and motion artifacts \cite{zhao2022jamsnet, zhang2024maskfusionnet, zhai2025research}. Early approaches are mostly frame-wise rPPG approaches, which can process each frame independently \cite{HR-CNN,DeepPhys,TS-CAN,liu2021efficientphys}. However, despite their frame-wise inference capability, these approaches have become less favored because of their limited temporal modeling capability. Subsequent approaches employ video clip inputs with Conv3D, Transformer, or Mamba architectures \cite{PhysNet, rPPGNet, physformer, yu2023physformer++, zou2025rhythmformer, zou2025rhythmmamba} to capture temporal dependencies and achieve remarkable performance, but they require long video segments (e.g., 128–160 frames) for accurate pulse recovery. Consequently, they cannot predict rPPG frame by frame and are therefore categorized as clip-wise rPPG approaches.

The reliance on long video segments fundamentally limits the practicality of clip-wise rPPG approaches. In real-world deployment, clip-wise inference requires capturing 4–5 seconds of video at 30 frames per second (FPS) \cite{zou2025rhythmmamba, braun2024suboptimal}, introducing a video acquisition latency as shown in Fig. \ref{fig:computational cost}. During video acquisition, users must remain still to avoid motion artifacts. This constraint degrades user experience and limits the feasibility of real-time streaming inference on edge devices. Although recent efforts have focused on model compression and lightweight design, these improvements are negligible compared to the latency introduced by long video acquisition. To achieve truly practical rPPG measurement, frame-wise rPPG estimation is therefore the most desirable solution. Yet, frame-wise rPPG prediction poses intrinsic challenges. Physiological rhythms such as heart rate (0.75–2.5 Hz) exhibit long temporal periods spanning several seconds \cite{cheng2020remote}, making it difficult to capture temporal and periodic dynamics and to accurately localize physiologically relevant facial regions from an isolated frame, resulting in poor accuracy and unstable generalization \cite{niu2020video}.

To address these challenges, we propose StreamPPG, an accurate low-latency frame-wise rPPG prediction architecture for real-time streaming rPPG estimation. StreamPPG supports highly parallelized training over full-video input, enabling frame-wise inference and thereby eliminating the latency bottleneck of long-video acquisition. Inspired by learning using privileged information (LUPI) \cite{vapnik2009new, pechyony2010theory, vapnik2015learning}, we propose consistent privileged learning (CPL) strategy, which leverages ground-truth rPPG signals as privileged information to strengthen the model’s representation capability while enforcing consistent behavior between training and inference without privileged input. Our StreamPPG achieves competitive accuracy compared with clip-wise rPPG approaches while maintaining frame-wise inference efficiency. Experiments on multiple datasets show that StreamPPG provides a balanced accuracy--efficiency trade-off and supports practical real-time streaming physiological measurement on edge devices. Our main contributions are summarized as follows:
\begin{itemize}
    \item We propose StreamPPG, an accurate low-latency frame-wise rPPG architecture capable of streaming inference with high accuracy, effectively eliminating video acquisition latency required for video clip input and achieving real-time performance.
    \item We introduce a consistent privileged learning (CPL) strategy that leverages ground-truth rPPG signals as privileged information to refine visual features, enabling the model to extract spatial-temporal representations from frame-wise inputs. This effectively overcomes the limitation of prior frame-wise approaches that struggle to capture temporal cues within isolated frames.
    \item We design an adaptive temporal modeling module (ATMM) which consists of two components: the adaptive attention enhancement block (AAEB), which improves encoder’s response to pulse-related facial regions, and the temporal state space block (TSSB), which models long-range temporal dependencies across frames.
    \item Extensive experiments across several datasets demonstrate that our architecture achieves state-of-the-art accuracy with low computational cost.
\end{itemize}

\section{Related Work}
\label{sec:related work}
\textbf{Traditional rPPG Approaches.} Remote photoplethysmography (rPPG) measures physiological signals by analyzing light reflected from the skin. Traditional rPPG approaches primarily rely on handcrafted feature extraction and frequency analysis within predefined facial regions. Early approaches such as Green [36] and ICA [26] estimate pulse signals from color variations in selected facial regions, with ICA further separating latent source components. Subsequent works like CHROM \cite{Chrome}, PBV \cite{PBV}, and POS \cite{POS} leverage the spectral characteristics of skin reflection to improve robustness against illumination changes. Despite these refinements, traditional methods remain sensitive to facial motion, occlusion, and lighting variations, often resulting in unstable or distorted physiological signals.

\textbf{Frame-wise Deep-Learning rPPG Approaches.} With the advent of deep-learning, data-driven rPPG estimation has significantly improved robustness and generalization in unconstrained environments. Early deep-learning rPPG models, such as HR-CNN \cite{HR-CNN}, TS-CAN \cite{TS-CAN}, DeepPhys \cite{DeepPhys}, and EfficientPhys \cite{liu2021efficientphys}, are capable of operating in a frame-wise manner. These models typically employ 2D-CNNs and spatial-attention modules to extract spatial representations and estimate rPPG signals directly from individual frames or shallow temporal cues. Although computationally efficient, their lack of explicit temporal modeling and weak attention to physiologically relevant facial regions severely limit their performance.

\textbf{Clip-wise Deep-Learning rPPG Approaches.} The mainstream deep-learning approaches rely on video segments (typically 128–160 frames) during both training and inference to capture spatial-temporal dependencies explicitly. Representative works such as PhysNet \cite{PhysNet}, rPPGNet \cite{rPPGNet}, and PhysFormer \cite{physformer} employ 3D convolutions or transformer-based architectures to model temporal dynamics and periodic relationships in physiological signals. JAMSNet \cite{zhao2022jamsnet} introduces joint attention and multi-scale fusion to enhance pulse extraction from different scale spaces. PhysFormer++ \cite{yu2023physformer++} further introduces a dual-branch fast–slow transformer to handle different temporal frequencies. TranPhys \cite{shao2023tranphys} adopts a spatiotemporal masked transformer to capture long-term contextual rhythm cues from facial video tubes. FactorizePhys \cite{joshi2024factorizephys} explores matrix-factorized multidimensional attention for rPPG representation learning. More recently, RhythmMamba \cite{zou2025rhythmmamba} leverages selective state-space modeling to achieve efficient long-range temporal reasoning with reduced computational cost. 

Nevertheless, their reliance on clip-wise inputs poses challenges for real-time or streaming applications, motivating our exploration of frame-wise rPPG architectures that jointly preserve temporal coherence and spatial attention on physiologically relevant facial regions.

\section{Method}
\label{sec:method}
Conventional clip-wise rPPG models rely on long video clips to capture temporal dependencies, introducing inevitable latency and limiting real-time applicability. To address this issue, StreamPPG is designed for accurate low-latency real-time streaming prediction using only frame-wise inputs while maintaining temporal awareness. Section \ref{subsec:streaming inference} first describes the real-time streaming inference pipeline that enables frame-wise, low-latency rPPG estimation. Section \ref{subsec:CPL strategy} then introduces the consistent privileged learning (CPL) strategy, which leverages privileged physiological signals during training to enhance representation learning and enforce consistency between training and inference. Finally, Section \ref{subsec:ATMM} presents the adaptive temporal modeling module (ATMM), which adaptively refines spatial and temporal features to preserve temporal coherence during frame-wise inference.

\subsection{Real-Time Streaming Inference Pipeline}
\label{subsec:streaming inference}

\begin{figure}[th!]
    \centering
    \includegraphics[width=0.9\columnwidth]{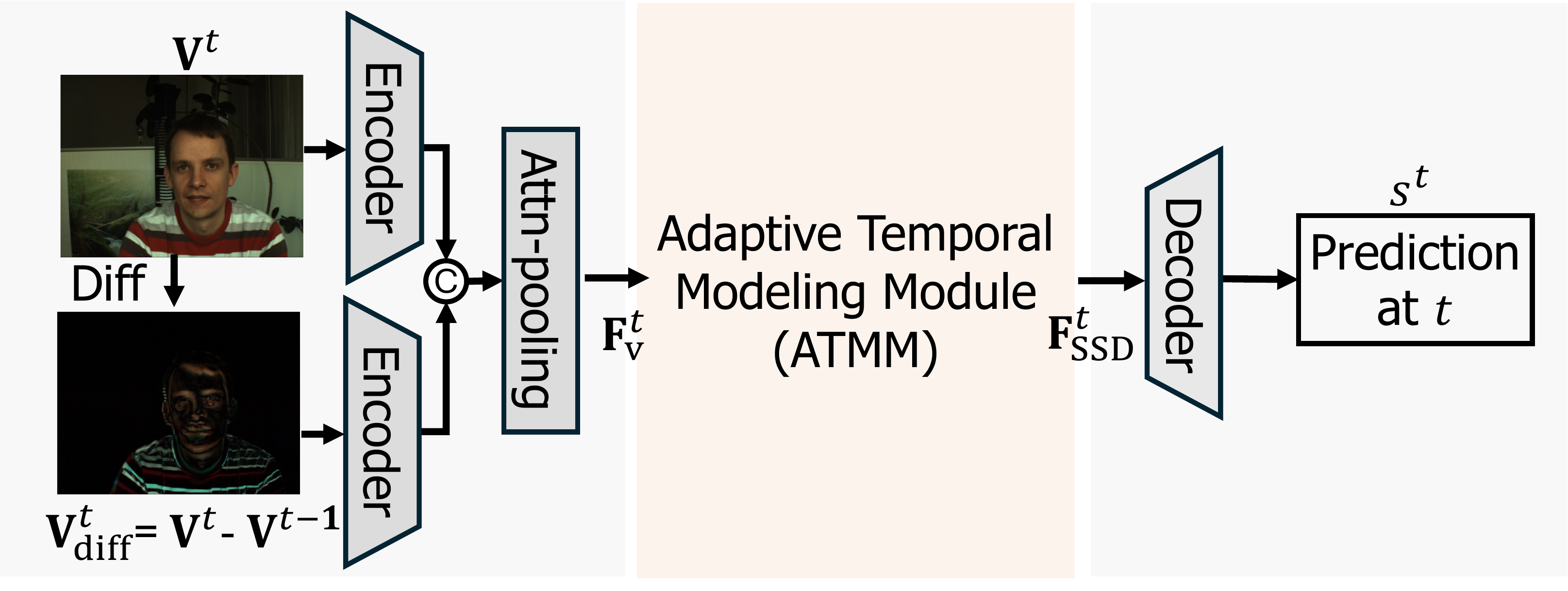}
    \caption{Overview of the StreamPPG inference pipeline. The model receives the current frame and its temporal difference as input, which are processed by dual image encoders followed by an adaptive temporal modeling module (ATMM) and a lightweight MLP decoder to predict the rPPG signal. StreamPPG performs real-time frame-wise streaming inference, enabling continuous physiological measurement without buffering long video segments.}
    \label{fig:Network_infer}
\end{figure}

StreamPPG performs accurate real-time streaming rPPG inference, eliminating the video acquisition latency imposed by video clip inputs. Operating in a frame-wise manner enables real-time streaming deployment on mobile and edge devices without clip buffering.

Fig. \ref{fig:Network_infer} illustrates the real-time streaming inference pipeline of StreamPPG. At time $t$, the model takes a frame $\mathbf{V}^t \in \mathbb{R}^{3\times H\times W}$. To enhance temporal sensitivity, a difference frame representation $\mathbf{V}_{\text{diff}}^t$ is constructed from consecutive frame differences. Two 2D-CNN encoders extract features from $\mathbf{V}^t$ and $\mathbf{V}_{\text{diff}}^t$, where the first encoder employs channel attention to emphasize physiologically relevant regions. The extracted features are concatenated to obtain the visual tokens $\mathbf{F}_{v}^t\in\mathbb{R}^{ D \times H' \times W'}$. ATMM then refines $\mathbf{F}_{v}^t$ to yield temporally enriched features $\mathbf{F}_{\text{SSD}}^t$. Finally, an MLP decoder produces the output signal $s^t\in\mathbb{R}$.

To further strengthen the model’s representational capability and make full use of physiological priors beyond what can be learned from frame-wise supervision, we introduce a consistent privileged learning (CPL) strategy. This training paradigm incorporates privileged physiological information available only during training to strengthen the model’s feature representations and ensure consistent behavior between training and inference.

\subsection{Consistent Privileged Learning Strategy}
\label{subsec:CPL strategy}

\begin{figure*}[th!]
    \centering
    \includegraphics[width=0.9\textwidth]{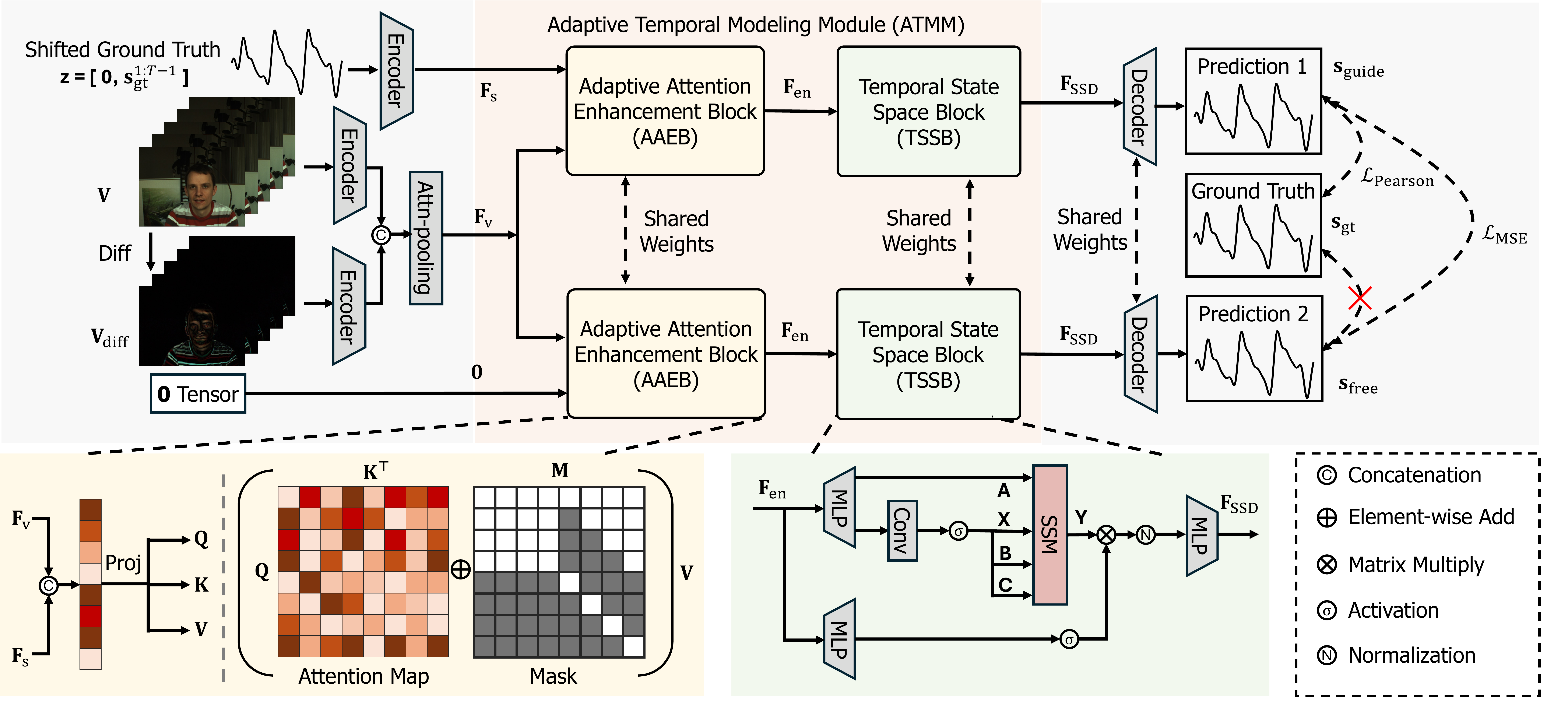}
    \caption{Overview of the StreamPPG training with consistent privileged learning (CPL) strategy. During training, privileged information is introduced through dual forward passes to enhance feature representation and enforce consistency between signal-guided and signal-free paths. The ATMM comprises two components, the adaptive attention enhancement block (AAEB) for emphasizing pulse-related facial regions and the temporal state space block (TSSB) for modeling long-range temporal dependencies.}
    \label{fig:Network_train}
\end{figure*}

We introduce a consistent privileged learning (CPL) strategy that leverages ground-truth rPPG signals as privileged information \cite{vapnik2009new, pechyony2010theory, vapnik2015learning} to enhance the model’s representation capability while ensuring consistency between training and inference without privileged inputs. 

Fig. \ref{fig:Network_train} shows an overview of the StreamPPG training with CPL strategy. During training, the model is optimized using entire video sequences instead of single frames, which allows efficient parallel computation and stable temporal learning while keeping the inference process strictly frame-wise. Concretely, each training iteration performs two forward passes under shared parameters $\theta$. The first, a signal-guided path, uses privileged information to produce $\mathbf{s}_{\text{guide}} = f_{\theta}(\mathbf{V}, \mathbf{z})$, with its expected risk $R_{\text{pr}}$ \cite{pechyony2010theory} defined as
\begin{equation}
R_{\text{pr}}(\theta) = \mathbb{E}\big[\mathcal{L}_{\text{Pearson}}(\mathbf{s}_{\text{guide}}, \mathbf{s}_{\text{gt}})\big], R_{\text{pr}}^* = \min_\theta R_{\text{pr}}(\theta).
\end{equation}
The second, a signal-free path, produces $\mathbf{s}_{\text{free}} = f_{\theta}(\mathbf{V}, \mathbf{0})$ without privileged input, and its expected risk $R_{\text{fr}}$ is given by
\begin{equation}
R_{\text{fr}}(\theta) = \mathbb{E}\big[\mathcal{L}_{\text{Pearson}}(\mathbf{s}_{\text{free}}, \mathbf{s}_{\text{gt}})\big], R_{\text{fr}}^* = \min_\theta R_{\text{fr}}(\theta).
\end{equation}
Meanwhile, introducing privileged information introduces a behavioral gap between the two forward paths, quantified as 
\begin{equation}
    \delta(\theta) = \mathbb{E}\!\left[\|f_\theta(\mathbf{V},\mathbf{z}) - f_\theta(\mathbf{V},\mathbf{0})\|_2\right].
\end{equation}
Assuming that the predictions are bounded after normalization and that the loss $\mathcal{L}_{\text{Pearson}}$ is $L$-Lipschitz in its first argument, the following bound holds (See Appendix A-B for detailed proof):
\begin{equation}
\label{eq:L-Lipschitz}
    R_{\text{fr}}(\theta) \le R_{\text{pr}}(\theta) + L \cdot \delta(\theta),
\end{equation}
where $L$ denotes the Lipschitz constant.

\textbf{Training Gains from Privileged Information.} Introducing privileged information provides additional information gain during training while also introducing a behavioral gap between the signal-guided and signal-free paths. To ensure that such learning remains beneficial, we analyze their quantitative relationship.

Let $\hat\theta_{\text{CPL}}$ be the parameter optimized under CPL, where $R_{\text{pr}}(\hat\theta_{\text{CPL}})\!\approx\!R_{\text{pr}}^*$. When the privileged signal carries information about the target (See Appendix A-C for detailed proof), $I(\mathbf{s}_{\text{gt}};\mathbf{z}\,|\,\mathbf{V})>0$,
\begin{equation}
\label{eq:info-gains}
\exists\, \epsilon_{\text{info}} > 0,\quad \text{ s.t. }\, R_{\text{fr}}^* - R_{\text{pr}}^* \ge \epsilon_{\text{info}}.
\end{equation}
Substituting $\hat\theta_{\text{CPL}}$ and $\epsilon_{\text{info}}$ into Eq. (\ref{eq:L-Lipschitz}) gives
\begin{equation}
R_{\text{fr}}(\hat\theta_{\text{CPL}}) \le R_{\text{fr}}^* + L\cdot\delta(\hat\theta_{\text{CPL}}) - \epsilon_{\text{info}}.
\end{equation}
Therefore, a sufficient condition for CPL to achieve lower inference risk than the baseline is $L\cdot\delta(\hat\theta_{\text{CPL}}) \le \epsilon_{\text{info}}$, which shows that CPL is theoretically advantageous when the consistency gap $\delta(\hat\theta_{\text{CPL}})$ is small and the information gain $\epsilon_{\text{info}}$ is large. 

Based on the above theoretical analysis, we design our framework to maximize the information gain $\epsilon_{\text{info}}$ by using shifted ground-truth rPPG signals as privileged input while maintaining temporal causality with an attention mask in the AAEB introduced in Section \ref{subsec:ATMM}. To minimize the consistency deviation $\delta(\hat{\theta}_{\text{CPL}})$, each training iteration performs two forward passes, one guided and one signal-free, which allows the model to explicitly quantify and reduce the discrepancy between them.

\textbf{Training-Inference Consistency Mechanism.} To enforce training-inference consistency under the dual-forward design, our formulation naturally yields three losses, $\mathcal{L}_{\text{Pearson}}\!\big(\mathbf{s}_{\text{guide}}, \mathbf{s}_{\text{gt}}\big)$, $\mathcal{L}_{\text{Pearson}}\!\big(\mathbf{s}_{\text{free}}, \mathbf{s}_{\text{gt}}\big)$ and $\mathcal{L}_{\text{MSE}}\!\big(\mathbf{s}_{\text{guide}}, \mathbf{s}_{\text{free}})$. We establish a generalization bound connecting the privileged and signal-free paths. The training objective combines two complementary goals, minimizing the privileged risk for supervised learning and constraining the prediction discrepancy between the two forward processes. Based on the following theoretical analysis, we derive the loss combination that minimizes the inference risk.

The empirical privileged risk $\hat{R}_{\text{pr}}(\theta)$ is defined as
\begin{equation}
    \hat{R}_{\text{pr}}(\theta) = \frac{1}{T} \sum_{t=1}^{T} \mathcal{L}_{\text{Pearson}}\!\big(f_\theta(\mathbf{V}^t, \mathbf{z}^{1:t}), s_{\text{gt}}^t\big),
\end{equation}
and the consistency regularization term $\hat{C}(\theta)$ measures the mean squared difference between the two forward paths,
\begin{equation}
    \hat{C}(\theta) = \frac{1}{T} \sum_{t=1}^{T}\!\big\|f_\theta(\mathbf{V}^{t}, \mathbf{z}^{1:t}) - f_\theta(\mathbf{V}^t, \mathbf{0})\big\|_2^2.
\end{equation}
We further employ the empirical Rademacher complexity \cite{bartlett2002rademacher} to characterize the generalization capacity of both the privileged predictor and the consistency function. Let $\mathfrak{R}_T(\mathcal{F}_{\text{pr}})$ and $\mathfrak{R}_T(\mathcal{G})$ denote their respective complexities. By applying the vector-contraction inequality \cite{maurer2016vector} and Eq. (\ref{eq:L-Lipschitz}), the following generalization bounds hold with probability at least $1-\gamma$ (See Appendix A-D for detailed proof):
\begin{equation}
\begin{aligned}
\label{eq:final risk bound}
R_{\text{fr}}(\theta)\le &\;\hat R_{\text{pr}}(\theta)+ 4L\,\mathfrak{R}_T(\mathcal{F}_{\text{pr}})+ 3\sqrt{\tfrac{\log(2/\gamma)}{2T}} \\
&\;+ L\,\sqrt{\hat C(\theta)+ 2\mathfrak{R}_T(\mathcal{G})+ 3\sqrt{\tfrac{\log(2/\gamma)}{2T}}}.
\end{aligned}
\end{equation}
This bound indicates that the inference risk $R_{\text{fr}}(\theta)$ can be minimized by jointly reducing the privileged loss and the prediction discrepancy between the two forward paths. Accordingly, the total optimization objective is defined as
\begin{equation}
\begin{aligned}
\mathcal{L}_{\text{total}} &= \mathcal{L}_{\text{Pearson}}\!\big(f_\theta(\mathbf{V},\mathbf{z}), \mathbf{s}_{\text{gt}}\big)\\
&\;+\lambda\,\mathcal{L}_{\text{MSE}}\!\big(f_\theta(\mathbf{V},\mathbf{z}), f_\theta(\mathbf{V},\mathbf{0})\big),
\end{aligned}
\end{equation}
where $\mathcal{L}_{\text{Pearson}}$ and $\mathcal{L}_{\text{MSE}}$ correspond to the empirical privileged risk $\hat R_{\text{pr
}}(\theta)$ and the consistency term $\hat C(\theta)$, respectively. The coefficient $\lambda$ controls the balance between supervision strength and consistency regularization.

In essence, this formulation provides a theoretical justification for the proposed consistency-regularized privileged learning: minimizing both losses jointly ensures the trained model achieves low inference risk even without privileged input.

\subsection{Adaptive Temporal Modeling Module}
\label{subsec:ATMM}
To further strengthen the model’s representation capability and ensure robust spatial-temporal awareness within frame-wise inference, we design the adaptive temporal modeling module (ATMM). This module jointly refines spatial encoding and temporal modeling, allowing the network to emphasize physiologically relevant regions while implicitly maintaining temporal coherence. Importantly, ATMM provides the architectural pathway for realizing CPL in a causal streaming setting, enabling privileged physiological guidance during training while preserving signal-free frame-wise inference at deployment.

\begin{figure}[th!]
    \centering
    \includegraphics[width=0.9\columnwidth]{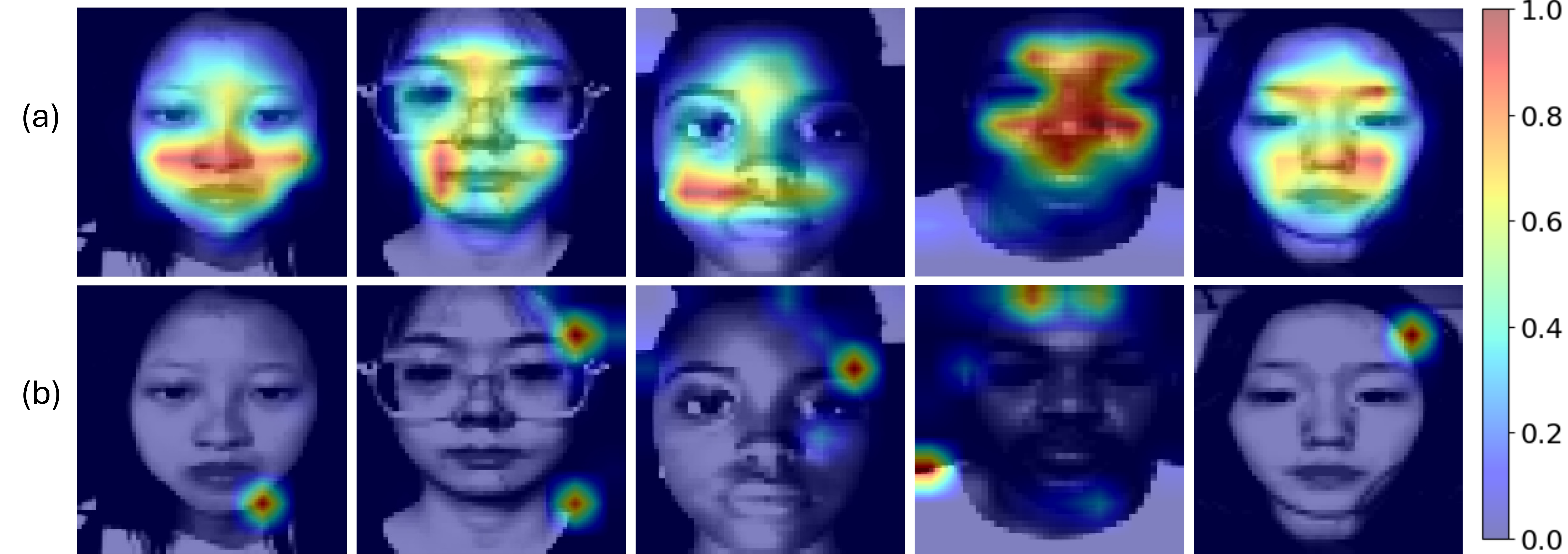}
    \caption{Visualization of facial attention maps. (a) The model trained with the AAEB. (b) The model trained without AAEB.}
    \label{fig:AAEB_attn}
\end{figure}

\textbf{Adaptive Attention Enhancement Block (AAEB)} is designed to incorporate historical rPPG signals as privileged information during training, enabling the model to learn how physiological dynamics influence visual representations. By leveraging this auxiliary signal pathway, AAEB adaptively refines the attention distribution within the visual encoder and the attention pooling, guiding the network to emphasize facial regions that relate to heart rate variation. Meanwhile, AAEB constrains the interaction between physiological and visual tokens to follow the temporal order, so that privileged guidance is introduced without violating the causality required by frame-wise inference. Fig. \ref{fig:AAEB_attn} visualizes the facial attention maps generated by models trained with and without the AAEB, demonstrating that AAEB strengthens focus on face regions most correlated with rPPG signals.

\begin{table*}[ht]
\caption{Intra-dataset evaluation results on several datasets. The best and second-best ones are in bold and underlined, respectively. ``F'' and ``C'' in the Type column denote frame-wise and clip-wise rPPG methods, respectively. ``-'' denotes results that are either not reported in the original paper or unavailable under the unified evaluation protocol.}
\label{tab:Intra_dataset_compare}
\tabcolsep=8pt
\centering
\renewcommand{\arraystretch}{1}
\resizebox{1.0\textwidth}{!}{
\begin{tabular}{lccccccccccccc}
\toprule
\multirow{3}{*}{Method} &\multirow{3}{*}{Type} &\multicolumn{3}{c}{PURE} &\multicolumn{3}{c}{UBFC} &\multicolumn{3}{c}{COHFACE} &\multicolumn{3}{c}{MMPD} \\
\cmidrule(lr){3-5} \cmidrule(lr){6-8} \cmidrule(lr){9-11} \cmidrule(lr){12-14}
 & &MAE &RMSE &$\rho$ &MAE &RMSE & $\rho$ &MAE &RMSE & $\rho$ &MAE &RMSE & $\rho$\\
\midrule
Face2PPG$_{\text{best}}$ \cite{casado2023face2ppg}  &F  &1.20 &1.80 &- &0.80 &1.10 &- &7.50  &9.80  &- &- &- &-    \\
DeepPhys \cite{DeepPhys} &F &0.83 & 1.54 & \textbf{0.99} & 6.27 & 10.82& 0.65  &6.89  &13.89  &0.34 &22.00 &28.13 &-0.07   \\
TS-CAN \cite{TS-CAN} &F &2.48 & 9.01 & \underline{0.92} & 1.70 & 2.72 & \textbf{0.99} &3.74 &9.68 &0.71 &12.03 &20.38 &0.17  \\
EfficientPhys \cite{liu2021efficientphys} &F & -    & -    & -    & 1.14 & 1.81 & \textbf{0.99} &31.27 &33.53 &0.11 &13.31 &20.73 &0.28  \\
PhysNet \cite{PhysNet}  &C &2.10 & 2.60 & \textbf{0.99} & 2.95 & 3.67 & 0.97 &5.38  &10.76  &- &4.93 &11.85 &0.62  \\
DeeprPPG \cite{liu2020general}  &C  &0.28 &0.43 &\textbf{0.99} &0.67 &1.70 &\textbf{0.99} &3.07  &7.06  &0.86 &- &- &-   \\
ETA-rPPGNet \cite{hu2021eta} &C &0.34 &0.77 &\textbf{0.99} &1.46 &3.97 &0.93 &4.67  &6.65  &0.77 &- &- &-  \\
PhysFormer \cite{physformer}  &C  & 1.10 & 1.75 & \textbf{0.99} & 0.50 & 0.71 & \textbf{0.99} &11.70  &13.83  &0.07 &11.99 &18.41 &0.18 \\
Li et al. \cite{li2023contactless} &C &0.64 &1.16 &\textbf{0.99} &0.48 &\underline{0.64} &\textbf{0.99} &-  &-  &- &- &- &-   \\
PFE-TFA \cite{li2023learning}  &C &- &- &- &- &- &- &1.31  &3.92  &- &- &- &-   \\
SINC \cite{speth2023non}  &C & 0.61 & 1.84 & \textbf{0.99}    &0.59  &1.83  &\textbf{0.99}  &- &- &-  &- &- &- \\
Contrast-Phys+ \cite{Contrast-Phys+} &C &0.48 &0.98 &\textbf{0.99} &\underline{0.21} &0.80 &\textbf{0.99} &- &- &- &- &- &- \\
RS+rPPG \cite{savic2025rs+} &C &0.34 &0.58 &\textbf{0.99} &0.62 &0.99 &\textbf{0.99} &- &- &- &- &- &- \\
MaskFusionNet \cite{zhang2024maskfusionnet} &C &1.11 &1.39 &\textbf{0.99} &- &- &- &\underline{1.27} &\underline{2.22} &\underline{0.98} &- &- &- \\
FactorizePhys \cite{joshi2024factorizephys} &C &3.08 &13.84 &0.91 &2.08 &3.63 &\underline{0.98} &5.25 &17.68 &0.53 &5.93 &12.74 &0.62 \\
RhythmMamba \cite{zou2025rhythmmamba} &C &\underline{0.23} &\underline{0.34} &\textbf{0.99} &0.50 &0.75 &\textbf{0.99} &- &- &- &\textbf{3.16} &\textbf{7.27} &\textbf{0.84}  \\
\rowcolor{gray!20}
\textbf{StreamPPG} &F & \textbf{0.20} & \textbf{0.33} & \textbf{0.99} & \textbf{0.14} & \textbf{0.27} & \textbf{0.99} &\textbf{0.71} &\textbf{1.97} &\textbf{0.99} &\underline{3.39} &\underline{8.35} &\underline{0.78} \\
\bottomrule
\end{tabular}
}
\end{table*}

Given the visual token $\mathbf{F}_{v}\in\mathbb{R}^{T\times D}$ and the encoded rPPG signal feature sequence $\mathbf{F}_{s}\in\mathbb{R}^{T\times D}$, we concatenate them along the temporal dimension to form the joint representation $\mathbf{H} = [\mathbf{F}_{s}, \mathbf{F}_{v}] \in \mathbb{R}^{2T\times D}$. Each token is further augmented with learnable modality and temporal embeddings before attention computation. To preserve causal alignment between visual and signal features during parallel training, while preventing future information leakage and maintaining temporal dependencies, we introduce an attention mask $\mathbf{M}$,
\begin{equation}
    \mathbf{M}(i,j) =
    \begin{cases}
        0, & \text{if } i,j \le T,\\[4pt]
        0, & \text{if } i > T \text{ and } j \le \min(T,\, i - T),\\[4pt]
        0, & \text{if } i = j,\\[4pt]
        -\infty, & \text{otherwise,}
    \end{cases}
\end{equation}
where the first case permits signal-to-signal attention, the second allows each visual token to attend only to past signal tokens, and the third enables self-attention. The first $T$ rows or columns correspond to signal tokens and the remaining $T$ correspond to visual tokens, respectively. The attention mask $\mathbf{M}$ is crucial for maintaining the causal structure during training, and without it, the model may learn unintended dependencies between visual and signal features. The attention output is computed as
\begin{equation}
    \mathrm{Attn}(\mathbf{F}_{s}, \mathbf{F}_{v}) = \mathrm{softmax}\!\left(\frac{\mathbf{Q}_{\text{attn}}\mathbf{K}_{\text{attn}}^\top}{\sqrt{D}} + \mathbf{M}\right)\mathbf{V}_{\text{attn}},
\end{equation}
where $\mathbf{Q}_{\text{attn}},\mathbf{K}_{\text{attn}},\mathbf{V}_{\text{attn}}$ are linear projections of $\mathbf{H}=[\mathbf{F}_{s}, \mathbf{F}_{v}]$. AAEB aligns visual tokens with physiological context and ensures causal and consistent behavior between training and inference.

\textbf{Temporal State Space Block (TSSB)} captures temporal dependencies of physiological dynamics across frames, implemented with a Mamba-2 state-space backbone \cite{dao2024transformers}. TSSB maintains an internal latent state that evolves recurrently over time, allowing the model to propagate rhythmic information and ensure temporal consistency during inference. The hidden state $h_t$ evolves according to an input-dependent transition,
\begin{equation}
\mathbf{h}^t = \mathbf{\bar{A}}^t \,\mathbf{h}^{t-1} + \mathbf{\bar{B}}^t \,\mathbf{F}_{\text{en}}^{t},
\end{equation}
where $\mathbf{F}_{\text{en}}^{t}$ is the output of AAEB, and $(\mathbf{\bar{A}}^t, \mathbf{\bar{B}}^t)$ are dynamically generated from the current feature $\mathbf{F}_{\text{en}}^{t}$. The temporally enhanced representation is then obtained as
\begin{equation}
    \mathbf{F}_{\text{SSD}}^{t} = \mathbf{\bar{C}}^t \,\mathbf{h}^t.
\end{equation}
The TSSB provides lightweight temporal state updates, helping StreamPPG maintain rhythmic continuity and temporal coherence during frame-wise inference. In our framework, TSSB serves as the temporal modeling component and works together with AAEB under the CPL strategy.

\section{Experiments}
\label{sec:exp}
\subsection{Datasets and Performance Metrics}
We conduct our experiments on four public datasets, the PURE \cite{puredataset}, UBFC-rPPG \cite{ubfcdataset}, COHFACE \cite{heusch2017reproducible}, and MMPD \cite{10340857}. The PURE dataset \cite{puredataset} contains 60 one-minute facial video sequences from 10 subjects, recorded under natural lighting. The subjects performed six head motion tasks. The videos have a resolution of 640$\times$480 pixels and a frame rate of 30 Hz. The UBFC-rPPG dataset \cite{ubfcdataset} contains 42 videos, each about 2 minutes long. The videos have a resolution of 640$\times$480 pixels and a frame rate of 30 Hz. The ground-truth bio-signals are captured using a pulse oximeter with a 60 Hz sampling rate. The COHFACE dataset \cite{heusch2017reproducible} has 160 video sequences, with each lasting one minute, collected from 40 subjects. The videos are recorded at a resolution of 640×480 pixels and 20 Hz using a Logitech HD C525 webcam. The corresponding blood volume pulse (BVP) and respiration signals are simultaneously recorded using Tought Technologies sensors and the BioGraph Infiniti software. The MMPD dataset \cite{10340857} contains 660 one-minute videos recorded using a Samsung Galaxy S22 Ultra, with a resolution of 320$\times$240 pixels at 30 FPS. It includes 33 participants with Fitzpatrick skin types 3-6, who perform four activities under four lighting conditions. Ground truth PPG signals are captured using a HKG-07C+ oximeter at 200 Hz, downsampled to 30 Hz.

\begin{figure}[b!]
    \centering
    \includegraphics[width=0.9\columnwidth]{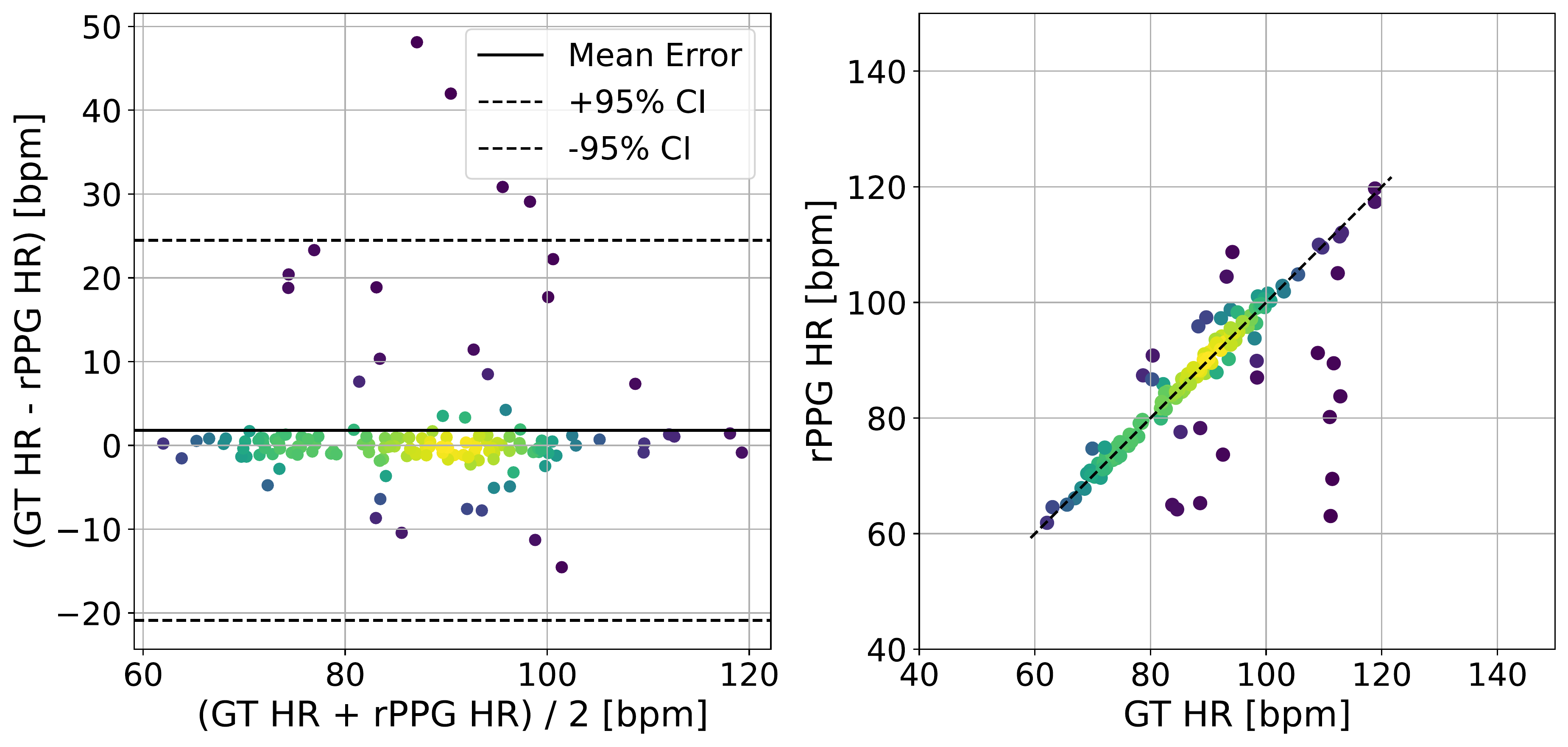}
    \caption{Bland-Altman plots and scatter plots from intra-dataset evaluation on MMPD. ``CI'' denotes the confidence interval.}
    \label{fig:BA_Image_Intra_MMPD}
\end{figure}

For evaluation, we use Mean Absolute Error (MAE), Root Mean Square Error (RMSE), Mean Absolute Percentage Error (MAPE), and Pearson Correlation Coefficient ($\rho$).

\begin{table*}[!th]
\caption{Cross-dataset evaluation results on several datasets. The first 7 rows are traditional rPPG methods, others are deep-learning methods. The best and second-best ones are in bold and underlined, respectively. ``F'' and ``C'' in the Type column denote frame-wise and clip-wise rPPG methods, respectively.}
\label{tab:Cross_dataset_compare}
\tabcolsep=4pt
\centering
\renewcommand{\arraystretch}{1}
\resizebox{1.0\textwidth}{!}{
\begin{tabular}{lc|cccccccccccccccc}
\toprule
& & \multicolumn{16}{c}{TrainSet $\rightarrow$ TestSet}  \\ 
\cmidrule{3-18} 
& & \multicolumn{4}{c|}{UBFC $\rightarrow$ PURE}   & \multicolumn{4}{c|}{PURE $\rightarrow$ UBFC}  & \multicolumn{4}{c|}{PURE $\rightarrow$ MMPD} & \multicolumn{4}{c}{UBFC $\rightarrow$ MMPD} \\ 
\midrule
Method  &Type & MAE & RMSE & MAPE & \multicolumn{1}{c|}{$\rho$} & MAE & RMSE & MAPE & \multicolumn{1}{c|}{$\rho$} & MAE & RMSE & MAPE & \multicolumn{1}{c|}{$\rho$} & MAE & RMSE & MAPE & $\rho$ \\ 
\midrule
Green \cite{Green} &F & 10.09  & 23.85  & 10.28 & \multicolumn{1}{c|}{0.34} & 19.73  & 31.00 & 18.72 & \multicolumn{1}{c|}{0.37} & 21.68 & 27.69 & 24.39 & \multicolumn{1}{c|}{-0.01} & 21.68 & 27.69 & 24.39 & -0.01 \\
ICA \cite{ICA} &F & 4.77 &16.07 & 4.47 &\multicolumn{1}{c|}{0.72} &16.00  &25.65  &15.35  &\multicolumn{1}{c|}{0.44} &18.60 &24.30 &20.88 &\multicolumn{1}{c|}{0.01} &18.60  &24.30 &20.88 &0.01      \\
CHROM \cite{Chrome} &F  &5.77 &14.93  &11.52  &\multicolumn{1}{c|}{0.81} &4.06 &8.83 &3.84 &\multicolumn{1}{c|}{0.89} &13.66 &18.76 &16.00 &\multicolumn{1}{c|}{0.08} &13.66  &18.76 &16.00 &0.08      \\
LGI \cite{LGI}  &F  &4.61 &15.38  &4.96 &\multicolumn{1}{c|}{0.77} &15.80  &28.55  &14.70  &\multicolumn{1}{c|}{0.36} &17.08 &23.32 &18.98 &\multicolumn{1}{c|}{0.04} &17.08  &23.32 &18.98 &0.04      \\
PBV \cite{PBV} &F   &3.92 &12.99  &4.84 &\multicolumn{1}{c|}{0.84} &15.90  &26.40  &15.17  &\multicolumn{1}{c|}{0.48} &17.95 &23.58 &20.18 &\multicolumn{1}{c|}{0.09} &17.95  &23.58 &20.18 &0.09      \\
POS \cite{POS}  &F  &3.67 &11.82  &7.25 &\multicolumn{1}{c|}{0.88} &4.08 &7.72 &3.93 &\multicolumn{1}{c|}{0.92} &12.36 &17.71 &14.43 &\multicolumn{1}{c|}{0.18} &12.36  &17.71 &14.43 &0.18      \\
OMIT \cite{casado2023face2ppg} &F &4.69 &15.82  &5.03 &\multicolumn{1}{c|}{0.76} &15.96  &28.57  &14.94  &\multicolumn{1}{c|}{0.36} &17.18 &23.07 &19.23 &\multicolumn{1}{c|}{0.07} &17.18  &23.07 &19.23 &0.07      \\ 
\midrule
DeepPhys \cite{DeepPhys} &F &5.54 &18.51  &5.32 &\multicolumn{1}{c|}{0.66} &1.21 &2.90 &1.42 &\multicolumn{1}{c|}{\textbf{0.99}}    & 16.92 &24.61 &18.54 &\multicolumn{1}{c|}{0.05} &17.50  &25.00 &19.27 &0.06      \\
TS-CAN \cite{TS-CAN} &F &3.69 &13.80  &\underline{3.39} & \multicolumn{1}{c|}{0.82} &1.30 &2.87 &1.50 &\multicolumn{1}{c|}{\textbf{0.99}}    & 13.94 &21.61 &15.15 &\multicolumn{1}{c|}{0.20} &14.01  &21.04 &15.48 &0.24      \\
EfficientPhys \cite{liu2021efficientphys} &F & 5.47 &17.04  &5.40 &\multicolumn{1}{c|}{0.71} &2.07 &6.32 &2.10 &\multicolumn{1}{c|}{0.94} &14.03 &21.62 &15.32 &\multicolumn{1}{c|}{0.17} &13.78  &22.25 &15.15 &0.09      \\
PhysNet \cite{PhysNet} &C &8.06 &19.71  &13.67  &\multicolumn{1}{c|}{0.61} &0.98 &2.48 &1.12 &\multicolumn{1}{c|}{\textbf{0.99}}    & 13.93 &20.29 &15.61 &\multicolumn{1}{c|}{0.17} &\underline{9.47} & \textbf{16.01}    & \underline{11.11} & 0.31      \\
PhysFormer \cite{physformer}  &C  & 12.92  &24.36  &23.92  &\multicolumn{1}{c|}{0.47} &1.44 &3.77 &1.66 &\multicolumn{1}{c|}{\underline{0.98}} & 14.57 &20.71 &16.73 &\multicolumn{1}{c|}{0.15} &12.10  &17.79 &15.41 &0.71      \\
Spiking-Phys \cite{liu2025spiking}  &C & 3.83 &-    &5.70 &\multicolumn{1}{c|}{0.83} &2.80 &-    &2.81 &\multicolumn{1}{c|}{0.95} &14.57 &-     &16.55 &\multicolumn{1}{c|}{0.14} &14.15  &-     &16.22 &0.15      \\
RhythmMamba \cite{zou2025rhythmmamba} &C  & \textbf{1.98}    & \textbf{6.51}    & 3.59 &\multicolumn{1}{c|}{\textbf{0.96}}    & \underline{0.95} & \underline{1.83} & \underline{1.04} & \multicolumn{1}{c|}{\textbf{0.99}}    & \underline{10.44} & \textbf{16.70}    & \underline{12.25} & \multicolumn{1}{c|}{\textbf{0.36}}    & 10.63  &17.14 &12.14 &\underline{0.34} \\
\rowcolor{gray!20}
\textbf{StreamPPG} &F  &\underline{2.00} & \underline{7.93} & \textbf{2.40}    & \multicolumn{1}{c|}{\underline{0.94}} & \textbf{0.33}    & \textbf{0.50}    & \textbf{0.32}    & \multicolumn{1}{c|}{\textbf{0.99}}    & \textbf{10.42}    & \underline{17.59} & \textbf{11.41}    & \multicolumn{1}{c|}{\underline{0.33}} & \textbf{9.29}    & \underline{16.46} & \textbf{10.28}    & \textbf{0.39}    \\ 
\bottomrule
\end{tabular}
}
\end{table*}

\subsection{Implementation Details}
Our method is implemented using PyTorch 2.4.0. We set the consistency loss weight to $\lambda=0.8$. We employ the AdamW optimizer with an initial learning rate $2\times10^{-4}$ and weight decay $1\times10^{-4}$ for training. We adopt a CosineAnnealing learning rate schedule with a linear warm-up for the first 3 epochs, a warm-up ratio of 0.1, and a minimum learning rate ratio of $1\times 10^{-2}$. The learning rate is updated per iteration, followed by cosine annealing over the remaining 27 epochs. We train the model for 30 epochs with a batch size of 8. All experiments are performed on an NVIDIA RTX A6000 GPU.

For all methods, we adopt the same preprocessing and postprocessing steps, and the same training protocols. In the preprocessing, video inputs are divided into segments of 128 frames. These segments are used only for training batching, while inference is performed in a frame-wise streaming manner. We crop and resize the face region to $128 \times128$ pixels. In the postprocessing, a second-order Butterworth filter with cutoff frequencies of 0.75--2.5 Hz is applied to filter the rPPG waveform. We use Welch algorithm to compute the power spectral density.

\begin{figure}[b!]
    \centering
    \includegraphics[width=0.9\columnwidth]{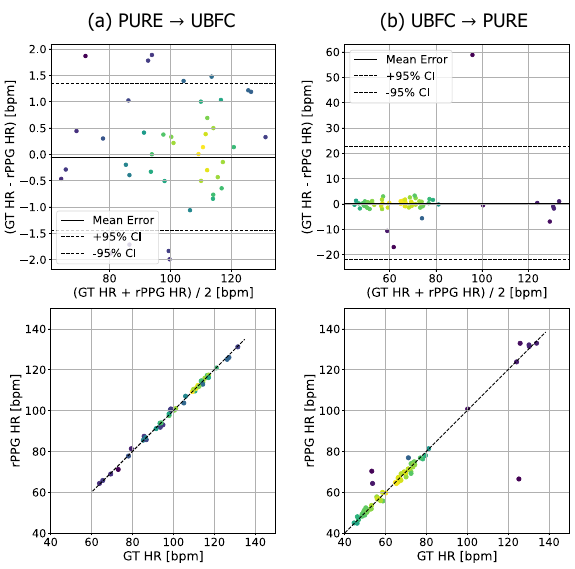}
    \caption{Bland-Altman plots and scatter plots. (a) Training with PURE and testing with UBFC, (b) training with UBFC and testing with PURE. ``CI'' denotes the confidence interval.}
    \label{fig:BA_Image_Cross}
\end{figure}

\subsection{Comparison Results}
We evaluate StreamPPG through both intra-dataset and cross-dataset experiments.

\begin{table}[ht]
    \centering
    \caption{Ablation of the CPL strategy under different motion conditions on MMPD.}
    \label{tab:ablation on CPL under motion}
    \centering
    \renewcommand{\arraystretch}{1}
    \tabcolsep=10pt
    \resizebox{\linewidth}{!}{
        \begin{tabular}{cc|cccc}
        \toprule
        \multicolumn{2}{c|}{Setting}   & MAE   & RMSE  & MAPE & $\rho$  \\ 
        \midrule
        \multicolumn{1}{c|}{\multirow{4}{*}{\begin{tabular}[c]{@{}c@{}}w/o\\ CPL\\ Strategy\end{tabular}}} & Stationary & 1.86  & 4.50  & 1.98 &0.92 \\
        \multicolumn{1}{c|}{}  & Rotation    & 4.39  & 7.89  & 5.99  & 0.64 \\
        \multicolumn{1}{c|}{}  & Talking     & 2.62  & 6.06  & 2.99  & 0.81 \\
        \multicolumn{1}{c|}{}  & Walking     & 10.97 & 17.14 & 9.78  & 0.03 \\ 
        \multicolumn{1}{c|}{}  & All Types   & 4.34  & 9.31  & 4.54  & 0.74 \\ 
        \midrule
        \multicolumn{1}{c|}{\multirow{4}{*}{\begin{tabular}[c]{@{}c@{}}w/\\ CPL\\ Strategy\end{tabular}}}  & Stationary & 0.90  & 2.01  & 1.02 &0.99 \\
        \multicolumn{1}{c|}{}  & Rotation    & 3.18  & 6.47  & 3.68  & 0.76 \\
        \multicolumn{1}{c|}{}  & Talking     & 1.77  & 5.15  & 2.08  & 0.87 \\
        \multicolumn{1}{c|}{}  & Walking     & 10.20 & 16.49 & 9.71  & 0.04 \\ 
        \multicolumn{1}{c|}{}  & All Types   & 3.39  & 8.35  & 3.50  & 0.78 \\ 
        \bottomrule
        \end{tabular}
    }
\end{table}
\begin{table}[ht]
    \centering
    \caption{Ablation of the CPL strategy under different illumination conditions on MMPD.}
    \label{tab:ablation on CPL under light}
    \centering
    \renewcommand{\arraystretch}{1}
    \tabcolsep=10pt
    \resizebox{\linewidth}{!}{
        \begin{tabular}{cc|cccc}
        \toprule
        \multicolumn{2}{c|}{Setting}   & MAE   & RMSE  & MAPE & $\rho$  \\ 
        \midrule
        \multicolumn{1}{c|}{\multirow{4}{*}{\begin{tabular}[c]{@{}c@{}}w/o\\ CPL\\ Strategy\end{tabular}}} & 100 lm & 4.26  & 7.87  & 4.65 & 0.79 \\
        \multicolumn{1}{c|}{}  & 200 lm      & 2.74  & 5.97  & 2.83  & 0.89 \\
        \multicolumn{1}{c|}{}  & 300 lm      & 4.76  & 10.35  & 5.16  & 0.75 \\
        \multicolumn{1}{c|}{}  & 300-800 lm  & 5.59  & 12.09 & 5.57  & 0.56 \\ 
        \multicolumn{1}{c|}{}  & All Types   & 4.34  & 9.31  & 4.54  & 0.74 \\ 
        \midrule
        \multicolumn{1}{c|}{\multirow{4}{*}{\begin{tabular}[c]{@{}c@{}}w/\\ CPL\\ Strategy\end{tabular}}}  & 100 lm & 3.35  & 7.04  & 3.84 & 0.84 \\
        \multicolumn{1}{c|}{}  & 200 lm      & 2.58  & 5.78  & 2.67  & 0.88 \\
        \multicolumn{1}{c|}{}  & 300 lm      & 3.67  & 8.16  & 3.87  & 0.82 \\
        \multicolumn{1}{c|}{}  & 300-800 lm  & 3.78  & 11.37 & 3.63  & 0.60 \\ 
        \multicolumn{1}{c|}{}  & All Types   & 3.39  & 8.35  & 3.50  & 0.78 \\ 
        \bottomrule
        \end{tabular}
    }
\end{table}
\begin{figure*}[ht!]
        \centering
        \includegraphics[width=1\linewidth]{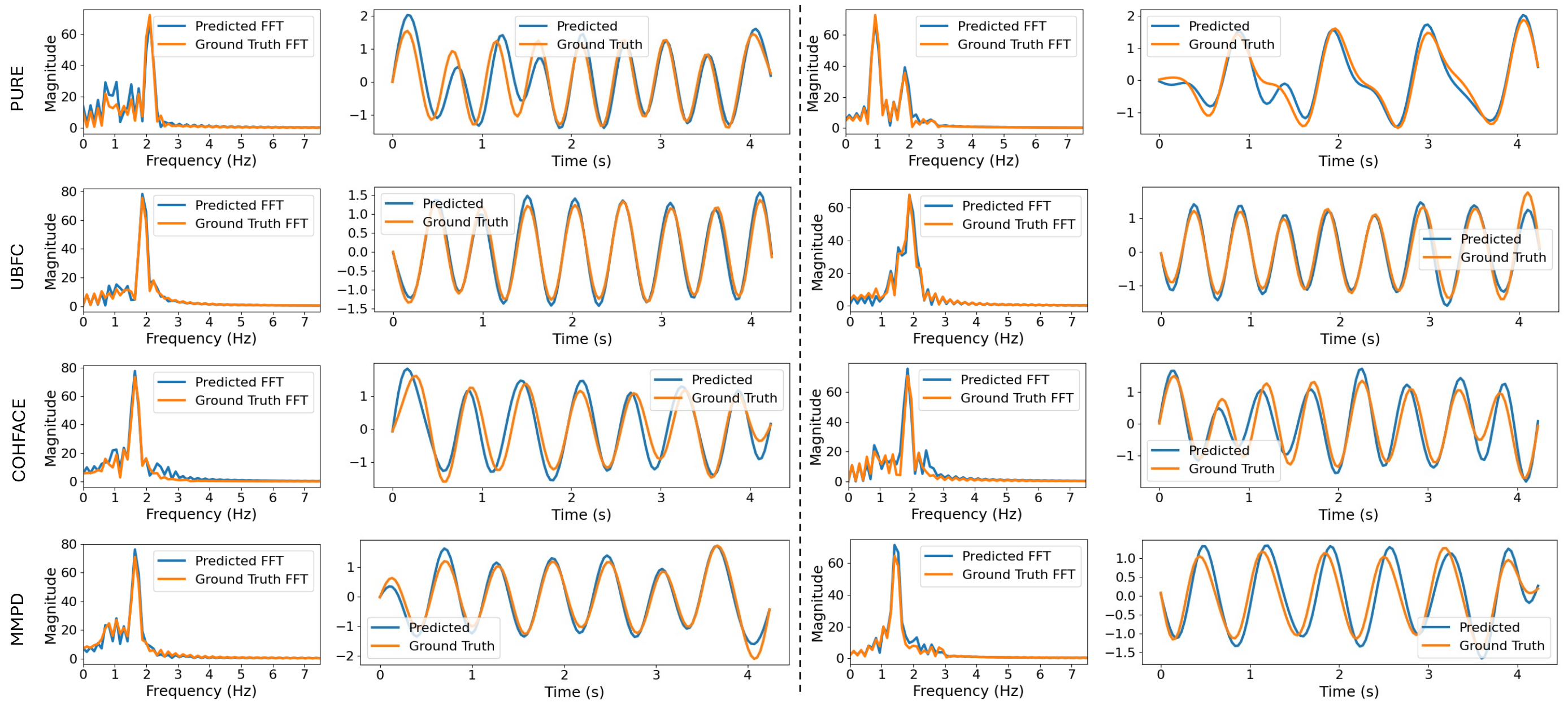}
        \caption{Representative rPPG signal visualizations from intra-dataset evaluation on the PURE, UBFC, COHFACE and MMPD datasets.}
        \label{fig:time-freq plot}
\end{figure*}

\textbf{Intra-Dataset Evaluation.} We conduct intra-dataset evaluations on four benchmark datasets. For the PURE dataset, we follow the protocol of \cite{lu2021dualgan, song2021pulsegan}, using the first 60\% of the data for training and the remaining 40\% for testing. On the UBFC dataset, we also follow \cite{lu2021dualgan, song2021pulsegan}, training on the first 30 subjects and testing on the remaining 12. For the COHFACE dataset, we adopt the setting in \cite{heusch2017reproducible}, where 60\% of the samples are used for training and the rest for testing. On the MMPD dataset, 70\% of the data is used for training, 10\% for validation, and 20\% for testing. Table \ref{tab:Intra_dataset_compare} shows that StreamPPG achieves state-of-the-art performance across all metrics on PURE, UBFC, and COHFACE, and delivers comparable results on MMPD compared to clip-wise rPPG approaches. Fig. \ref{fig:time-freq plot} presents representative examples of the predicted and ground-truth signals from the PURE, UBFC, COHFACE, and MMPD datasets. For visual clarity, all signals are bandpass-filtered between 0.75 Hz and 2.5 Hz to isolate the HR-related components. The left column shows the frequency spectra, where the dominant peaks of the predicted signals are generally aligned with those of the ground truth. The right column displays the corresponding time-domain waveforms, where the predicted rPPG signals follow the main periodic patterns of the ground-truth PPG. These visualizations indicate that the proposed model captures both frequency-domain HR information and temporally coherent waveform patterns. Fig. \ref{fig:BA_Image_Intra_MMPD} shows that our StreamPPG achieves low bias and strong correlation with the ground-truth rPPG signals.

\textbf{Cross-Dataset Evaluation.} For a fair comparison, we conduct our cross-dataset evaluation experiments on PURE, UBFC and MMPD datasets following the protocol outlined in \cite{liu2022rppg}. We train the models on PURE and UBFC individually and then test on PURE, UBFC and MMPD datasets. The training dataset is split into a training set and a validation set, with the first 80\% used for training and the remaining 20\% used for validation. Table \ref{tab:Cross_dataset_compare} shows that our StreamPPG consistently achieves the best or second-best results across datasets. Fig. \ref{fig:BA_Image_Cross} further shows the cross-dataset evaluation results between the PURE and UBFC datasets. When trained on PURE and tested on UBFC, the Bland-Altman plot shows a very narrow 95\% confidence interval (CI), and the scatter points align almost perfectly along the diagonal, indicating strong generalization and consistency. This can be attributed to the higher diversity of the PURE dataset, which provides more robust representations transferable to UBFC. Conversely, when trained on UBFC and tested on PURE, the 95\% CI becomes noticeably wider, and several points in the scatter plot deviate from the diagonal, although most predictions remain accurate. This asymmetry highlights that models trained on smaller datasets may not fully capture the domain variability required for robust generalization across datasets.

\begin{table}[ht]
\caption{Ablation study on different loss combinations on the MMPD dataset. $\mathbf{s}_{\text{gt}}$, $\mathbf{s}_{\text{guide}}=f_\theta(\mathbf{V},\mathbf{Z})$, $\mathbf{s}_{\text{free}}=f_\theta(\mathbf{V},\mathbf{0})$ denote the ground-truth signal, prediction with privileged information and the prediction signal without it, respectively.}
\label{tab:ablation on Loss}
\centering
\renewcommand{\arraystretch}{1}
\tabcolsep=4pt
\resizebox{1.0\columnwidth}{!}{
\begin{tabular}{c|ccc|cccc}
\toprule
\multirow{3}{*}{Setting} &\multicolumn{3}{c|}{Loss} & \multicolumn{4}{c}{MMPD Dataset} \\
\cmidrule{2-8}
&$\mathcal{L}_{\text{MSE}}$ &$\mathcal{L}_{\text{Pearson}}$ &$\mathcal{L}_{\text{Pearson}}$ & \multirow{2}{*}{MAE}    & \multirow{2}{*}{RMSE}   & \multirow{2}{*}{MAPE}   &\multirow{2}{*}{$\rho$}   \\
&$(\mathbf{s}_{\text{guide}},\mathbf{s}_{\text{free}})$ &$(\mathbf{s}_{\text{guide}},\mathbf{s}_{\text{gt}})$ &$(\mathbf{s}_{\text{free}},\mathbf{s}_{\text{gt}})$ & & & & \\
\midrule
1 &\checkmark &\checkmark &\checkmark &4.97    &10.38   &5.09    &0.68     \\
2 &           &\checkmark &\checkmark &5.74    &11.89   &5.92    &0.60   \\
3 &\checkmark &           &\checkmark &3.96    &8.39    &4.02    &0.77   \\
4 &           &\checkmark &           &9.58    &13.19   &10.42   &0.31   \\
5 &           &           &\checkmark &3.98    &8.90    &4.18    &0.76   \\
\rowcolor{gray!20}
6 &\checkmark &\checkmark &           &3.39    &8.35    &3.50    &0.78    \\
\bottomrule
\end{tabular}
}
\end{table}

\subsection{Ablation Study}
We conduct ablation studies to evaluate the contribution of StreamPPG. 

\textbf{Ablation on the CPL Strategy.} We first evaluate the effectiveness of the proposed CPL strategy under different motion and illumination conditions. Table \ref{tab:ablation on CPL under motion} and Table \ref{tab:ablation on CPL under light} show that CPL consistently improves all metrics across all motion types and illumination types. These empirical results are consistent with the intuition behind Eq. (\ref{eq:info-gains}), suggesting that privileged physiological information provides useful guidance under challenging conditions.

\begin{figure}[b!]
        \centering
        \includegraphics[width=1.0\columnwidth]{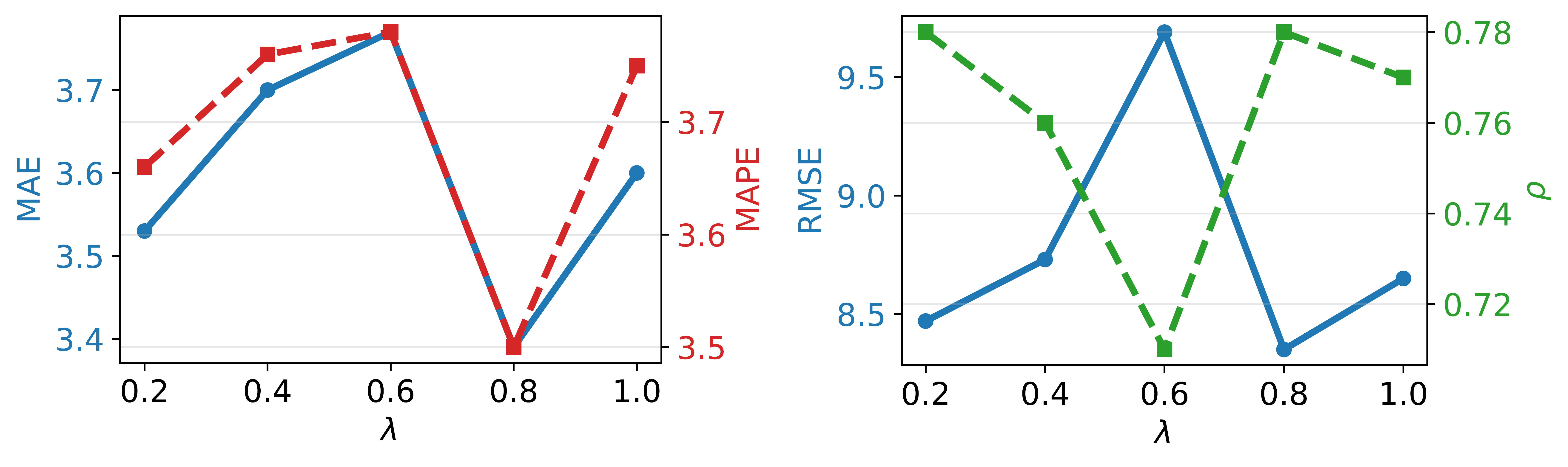}
        \caption{Effect of $\lambda$ on the MMPD dataset. Performance peaks at $\lambda=0.8$, which provides the best balance between privileged supervision and consistency regularization.}
        \label{fig:hyperparam}
\end{figure}
\begin{figure}[b!]
        \centering
        \includegraphics[width=0.93\columnwidth]{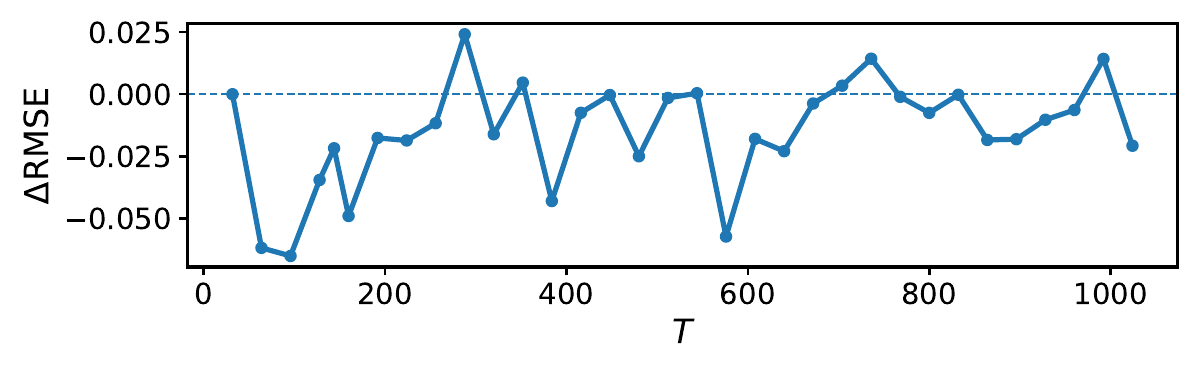}
        \caption{Relative change of waveform RMSE versus inference length on MMPD under frame-wise streaming inference. $\Delta$RMSE is calculated as $\mathrm{RMSE}(T)-\mathrm{RMSE}(T_{\mathrm{ref}})$, where $T_{\mathrm{ref}}$ denotes 32 frames.}
        \label{fig:rmse_rppg}
\end{figure}

\begin{table}[b!]
\caption{Ablation study on the encoder of StreamPPG on the MMPD dataset. ``CA'' denotes channel attention.}
\label{tab:ablation on Encoder}
\centering
\footnotesize
\renewcommand{\arraystretch}{1}
\tabcolsep=18pt
\resizebox{1.0\columnwidth}{!}{
    \begin{tabular}{c|cccc}
    \toprule
    Encoder & MAE    & RMSE   & MAPE   &$\rho$ \\
    \midrule
    w/o CA               &3.77    &9.35    &3.92    &0.74   \\
    \rowcolor{gray!20}
     w/ CA       &3.39    &8.35    &3.50    &0.78   \\  
    \bottomrule
    \end{tabular}
}
\end{table}

Table \ref{tab:ablation on Loss} shows that training only with privileged supervision (setting 4) causes severe overfitting to the signal-guided path and poor generalization in signal-free inference, confirming the existence of the training–inference gap. Using only the signal-free loss (setting 5) or combining it with the signal-guided loss (setting 2) still yields weak performance, as the gap remains uncontrolled and spatial-temporal features are insufficiently guided. Applying only the consistency and signal-free losses (setting 3) alleviates the gap but lacks privileged information to strengthen representation learning, resulting in limited accuracy. Jointly optimizing all three losses (setting 1) introduces interference between the two forward paths, leading to unstable optimization. The best results arise from combining privileged supervision with consistency regularization (setting 6), where the signal-guided forward enhances representational learning and the consistency term mitigates the training–inference discrepancy, providing empirical support for the proposed loss formulation derived from Eq. (\ref{eq:final risk bound}).

We further conduct an ablation study over $\lambda$. Fig. \ref{fig:hyperparam} illustrates that $\lambda=0.8$ achieves the best balance between privileged supervision and the consistency regularization term, resulting in the lowest overall error.

\textbf{Ablation on Image Encoder.} We also conduct an ablation study on the channel attention in the image encoder. Table \ref{tab:ablation on Encoder} shows that removing the channel-attention module leads to clear performance degradation, indicating that the channel-attention mechanism enhances the encoder’s representation capacity.

\begin{table}[b!]
    \centering
    \caption{Ablation study on attention mask $\textbf{M}$ in AAEB on MMPD. ``AM'' denotes our designed attention mask.}
    \label{tab:ablation on mask in AAEB}
    \centering
    \footnotesize
    \renewcommand{\arraystretch}{1}
    \tabcolsep=18pt
    \resizebox{\columnwidth}{!}{
    \begin{tabular}{c|cccc}
    \toprule
    \multirow{2}{*}{AAEB} &\multicolumn{4}{c}{MMPD Dataset} \\
    \cmidrule{2-5}
     &MAE     &RMSE     &MAPE    &$\rho$       \\
    \midrule
    w/o AM   &10.50        &15.95        &10.97        &0.33        \\
    \rowcolor{gray!20}
    w/ AM    &3.39    &8.35   &3.50     &0.78       \\
    \bottomrule
    \end{tabular}
    }
\end{table}
\begin{table}[b!]
        \caption{Ablation study on the ATMM of StreamPPG on MMPD. Starting from the full model, TSSB is first replaced with an LSTM-based temporal module with a comparable parameter scale. AAEB and CPL are then progressively removed to evaluate the contribution of each component.}
        \label{tab:ablation on ATMM}
        \centering
        \footnotesize
        \renewcommand{\arraystretch}{1.0}
        \tabcolsep=11pt
        \resizebox{\columnwidth}{!}{
        \begin{tabular}{c|cc|cccc}
        \toprule
        \multirow{2}{*}{CPL}    &\multicolumn{2}{c|}{ATMM}  &\multicolumn{4}{c}{MMPD Dataset} \\
        \cmidrule{2-3}  \cmidrule{4-7}
                      & AAEB        & TSSB        &MAE     &RMSE    &MAPE    &$\rho$     \\ 
        \midrule
                      &             &\checkmark   &4.34    &9.31    &4.54    &0.74       \\
        \checkmark    &             &\checkmark   &3.98    &8.90    &4.18    &0.76       \\
        \checkmark    &\checkmark   &             &4.07    &8.59    &4.06    &0.76       \\
        \rowcolor{gray!20}
        \checkmark    &\checkmark   &\checkmark   &3.39    &8.35    &3.50    &0.78       \\
        \bottomrule
        \end{tabular}
        }
\end{table}

\textbf{Ablation on ATMM.} We further perform an ablation study to evaluate the contribution of each component in ATMM. Table \ref{tab:ablation on mask in AAEB} shows that removing the adaptive mask from AAEB results in significant performance degradation across all metrics. Without the mask, the attention mechanism can focus on future privileged signals, introducing unintended dependencies. This violates the causal consistency required by frame-wise inference. At test time, privileged signals are unavailable, which introduces a train--test mismatch and leads to a substantial performance drop. Table \ref{tab:ablation on ATMM} further shows that removing the AAEB leads to clear performance degradation across all metrics. This indicates that AAEB leverages privileged signal guidance to refine spatial attention and strengthen spatiotemporal representations. We further replace TSSB with a standard LSTM-based temporal module and observe a performance drop, indicating that TSSB is important for preserving physiological dynamics across frames. Fig. \ref{fig:rmse_rppg} further examines long-sequence stability by showing the RMSE between the predicted and ground-truth rPPG signals (i.e., waveform RMSE rather than HR error) as the inferred sequence length increases. The error remains within a narrow range without an evident upward trend, indicating no obvious cumulative drift during long-sequence prediction. This supports the stability of the proposed temporal-state mechanism under frame-wise streaming inference.

\textbf{Discussion.} The central contribution of our work is a novel accurate frame-wise inference architecture for rPPG estimation, with the key innovation being the CPL strategy. As shown in Table \ref{tab:ablation on CPL under motion} and Table \ref{tab:ablation on ATMM}, CPL is the key enabler of the improvements by leveraging privileged supervision while controlling the training–inference gap. ATMM serves as an architectural realization of the streaming pipeline, offering complementary spatial and temporal modeling ability rather than being the main source of gains.

\begin{table}[ht]
\caption{Computational cost and inference efficiency. RMSE is tested on UBFC dataset. ``F'' and ``C'' in the Type column denote frame-wise and clip-wise rPPG methods, respectively.}
\label{tab:computation}
\centering
\renewcommand{\arraystretch}{1}
\tabcolsep=5.5pt
\resizebox{1.0\columnwidth}{!}{
\begin{tabular}{lc|ccccc}
\toprule
\footnotesize
\multirow{2}{*}{Method} &\multirow{2}{*}{Type} &Param. &MACs &Throughput &Memory &RMSE \\
 &  &(M) &(G) &(FPS) &(MB) &(bpm) \\
\midrule
DeepPhys &F &1.98 & 0.74 & 1798 & 178.04 &10.82\\
TS-CAN   &F &1.98 & 0.74 & 1356 & 186.60 &2.72\\
EfficientPhys  &F & 1.91 & 0.37 & 1550 & 124.61 &1.81\\
PhysNet &C &0.77 & 70.12  & 102  & 432.96 &3.67\\
PhysFormer    &C  & 7.38 & 50.61  & 80   & 598.53 &0.71\\
RhythmMamba  &C & 1.07 & 12.94  & 51   & 283.00 &0.75 \\
\rowcolor{gray!20}
StreamPPG &F & 2.92 & 1.59 & 290 & 405.75 &0.27 \\
\bottomrule
\end{tabular}
}
\end{table}
\begin{table}[ht]
  \centering
  \caption{End-to-end throughput and accuracy on the UBFC dataset using a Jetson AGX Orin under different input resolutions. ``Thr.'' denotes end-to-end throughput measured in FPS and includes preprocessing, model inference, and postprocessing. RMSE is reported in bpm.}
  \label{tab:end-to-end throughput}
  \renewcommand{\arraystretch}{1.0}
    \setlength{\tabcolsep}{4pt} 
    \resizebox{1.0\columnwidth}{!}{
    \begin{tabular}{c|cc|cc|cc|cc}
    \toprule
    \multirow{2}{*}{Method}  & \multicolumn{2}{c|}{64$\times$64} & \multicolumn{2}{c|}{96$\times$96} & \multicolumn{2}{c|}{128$\times$128} & \multicolumn{2}{c}{256$\times$256} \\ 
    \cmidrule{2-9} 
            & Thr.    & RMSE    & Thr.    & RMSE    & Thr.     & RMSE    & Thr.    & RMSE    \\
    \midrule
    TS-CAN       & 239  & 2.78  & 231  & 2.75  & 227            & 2.72    & 70            & 2.31    \\
    PhysNet      & 19   & 8.05   & 12            & 3.35    & 6              & 3.67    & 1             & 0.86    \\
    RhythmMamba            & 8       & 1.03    & 6        & 0.85    & 6         & 0.75    & 2             & 0.78    \\
    \rowcolor{gray!20}
    StreamPPG              & 56    & 0.31    & 54       & 0.27    & 53          & 0.27    & 54          & 0.22    \\ 
    \bottomrule
    \end{tabular}
    }
\end{table}

\subsection{Computational Efficiency}
To evaluate computational efficiency under a realistic streaming scenario, we adopt a sliding-window inference protocol rather than processing the entire video sequence in a single forward pass. This protocol reflects the deployment behavior of clip-wise models, which must wait until a full clip is available and then reprocess overlapping clips as new frames arrive. In contrast, frame-wise models update predictions sequentially with each incoming frame. The measurement protocol is summarized as follows:
\begin{itemize}
    \item Warm-up: Each model is first run for 50 forward passes before timing.
    \item Timing: Throughput is measured over a 1000-frame sequence. Clip-wise models use 160-frame sliding windows and perform 841 forward passes, while frame-wise models process incoming frames sequentially and perform 999 forward passes.
    \item Repetition and averaging: Each measurement is repeated 10 times, and the average throughput is reported.
\end{itemize}

Table \ref{tab:computation} compares the computational efficiency of different methods. StreamPPG provides a balanced trade-off between accuracy and efficiency, with relatively low computational cost and real-time throughput while maintaining accurate rPPG estimation. These results support the practicality of the proposed frame-wise streaming framework for low-latency rPPG inference.

We further report end-to-end throughput and accuracy on a Jetson AGX Orin as an edge-device deployment reference, covering the practical inference pipeline under different input resolutions. As shown in Table \ref{tab:end-to-end throughput}, StreamPPG achieves real-time throughput across all tested resolutions while achieving the lowest RMSE among the compared approaches. Although its raw throughput is lower than TS-CAN, StreamPPG provides a better accuracy-efficiency tradeoff and remains faster than clip-wise methods such as PhysNet and RhythmMamba. These results demonstrate that the proposed strict frame-wise streaming framework is practical for edge deployment while preserving high rPPG estimation accuracy.

\section{Conclusions}
\label{sec:conclusion}
In this work, we propose StreamPPG, an accurate low-latency frame-wise rPPG architecture capable of real-time streaming inference, achieving competitive accuracy compared with clip-wise rPPG approaches. Furthermore, we introduce the CPL strategy, which leverages ground-truth physiological signals as privileged information to enhance the model’s representation capability and enforce training–inference consistency through a theoretically justified loss formulation. We also design ATMM to enhance temporal awareness and spatial focus, enabling robust feature encoding under frame-by-frame inference. Extensive experiments across multiple datasets further demonstrate that StreamPPG achieves state-of-the-art performance with low computational cost.

\bibliographystyle{plain}   
\bibliography{reference} 

\appendices
\section{Consistent Privileged Learning Strategy}
\label{suppl:CPL proof}
\subsection{Notation and Definitions}
$\mathbf{V}$, $\mathbf{z}$ and $\mathbf{s}_{\text{gt}}$ denote the input video, privileged information and ground-truth rPPG signal, respectively.

During the training process, each iteration has two forward passes under our consistent privileged learning (CPL) strategy. The first, a signal-guided path, uses privileged information to produce $\mathbf{s}_{\text{guide}} = f_{\theta}(\mathbf{V}, \mathbf{z})$, with its expected risk $R_{\text{pr}}$ defined as
\begin{equation}
\begin{aligned}
R_{\text{pr}}(\theta) &= \mathbb{E}\big[\mathcal{L}_{\text{Pearson}}(\mathbf{s}_{\text{guide}}, \mathbf{s}_{\text{gt}})\big], \\
R_{\text{pr}}^* &= \min_\theta R_{\text{pr}}(\theta).
\end{aligned}
\end{equation}
The second, a signal-free path, produces $\mathbf{s}_{\text{free}} = f_{\theta}(\mathbf{V}, \mathbf{0})$ without privileged input, and its expected risk $R_{\text{fr}}$ is given by
\begin{equation}
\begin{aligned}
R_{\text{fr}}(\theta) &= \mathbb{E}\big[\mathcal{L}_{\text{Pearson}}(\mathbf{s}_{\text{free}}, \mathbf{s}_{\text{gt}})\big], \\
R_{\text{fr}}^* &= \min_\theta R_{\text{fr}}(\theta).  
\end{aligned}
\end{equation}
Meanwhile, introducing privileged information introduces a behavioral gap between the two forward paths, quantified as 
\begin{equation}
    \delta(\theta) = \mathbb{E}\left[\|f_\theta(\mathbf{V},\mathbf{z}) - f_\theta(\mathbf{V},\mathbf{0})\|_2\right].  
\end{equation}
The empirical privileged risk $\hat{R}_{\text{pr}}(\theta)$ is defined as
\begin{equation}
    \hat{R}_{\text{pr}}(\theta) = \frac{1}{T} \sum_{t=1}^{T} \mathcal{L}_{\text{Pearson}}\big(f_\theta(\mathbf{V}^t, \mathbf{z}^{1:t}), s_{\text{gt}}^t\big),  
\end{equation}
and the consistency regularization term $\hat{C}(\theta)$ measures the mean squared difference between the two forward paths,
\begin{equation}
    \hat{C}(\theta) = \frac{1}{T} \sum_{t=1}^{T}\big\|f_\theta(\mathbf{V}^{t}, \mathbf{z}^{1:t}) - f_\theta(\mathbf{V}^t, \mathbf{0})\big\|_2^2.  
\end{equation}

\subsection{Proof of Eq. (4)}
\label{appendix:proof of eq4}
The loss $\mathcal{L}_{\text{Pearson}}$ is $L$-Lipschitz, for any $\theta \in \Theta$,
\begin{equation}
    R_\text{fr}(\theta) \le R_{\text{pr}}(\theta) + L \cdot \delta(\theta).  
\end{equation}

\begin{proof}
By the $L$-Lipschitz continuity of $\mathcal{L}_{\text{Pearson}}$
\begin{equation}
\begin{aligned}
\mathcal{L}_{\text{Pearson}}(f_\theta(\mathbf{V}, \mathbf{0}), \mathbf{s}_{\text{gt}})&\le \mathcal{L}_{\text{Pearson}}(f_\theta(\mathbf{V}, \mathbf{z}), \mathbf{s}_{\text{gt}}) \\
&+ L \|f_\theta(\mathbf{V}, \mathbf{0}) - f_\theta(\mathbf{V}, \mathbf{z})\|_2.
\end{aligned}
\end{equation}
Taking expectations gives
\begin{equation}
\label{supply_eq:Lipschitz}
R_{\text{fr}}(\theta)\le  R_{\text{pr}}(\theta) + L \cdot \delta(\theta).
\end{equation}
\end{proof}

\subsection{Proof of Eq. (5)}
\label{appendix:proof of eq5}
When the privileged signal carries information about the target, $I(\mathbf{s}_{\text{gt}};\mathbf{z}\,|\,\mathbf{V})>0$,
\begin{equation}
\exists\, \epsilon_{\text{info}} > 0,\quad \text{ s.t. }\, R_{\text{fr}}^* - R_{\text{pr}}^* \ge \epsilon_{\text{info}}.  
\end{equation}

\begin{proof}
We use the neg-Pearson loss
\begin{equation}
\mathcal{L}_{\text{Pearson}}(\mathbf{a},\mathbf{b}) = 1 - \mathrm{Corr}(\mathbf{a},\mathbf{b}),
\end{equation}
where
\begin{equation}
\mathrm{Corr}(\mathbf{a},\mathbf{b}) = \frac{\mathrm{Cov}(\mathbf{a},\mathbf{b})}{\sqrt{\mathrm{Var}(\mathbf{a})\,\mathrm{Var}(\mathbf{b})}}.
\end{equation}
For any observable $O\in\{\mathbf{V},(\mathbf{V},\mathbf{z})\}$, define the correlation ratio
\begin{equation}
\eta^2(\mathbf{s}_{\text{gt}}\mid O) = \frac{\mathrm{Var}\big(\mathbb{E}[\mathbf{s}_{\text{gt}}\mid O]\big)}{\mathrm{Var}(\mathbf{s}_{\text{gt}})}\in[0,1].
\end{equation}
By the tower property and the Cauchy-Schwarz inequality, for any square-integrable $g(O)$ with zero mean and unit variance,
\begin{equation}
\mathrm{Corr}\big(\mathbf{s}_{\text{gt}},g(O)\big)
\le \frac{\sqrt{\mathrm{Var}(\mathbb{E}[\mathbf{s}_{\text{gt}}\mid O])}}{\sqrt{\mathrm{Var}(\mathbf{s}_{\text{gt}})}}
= \eta(\mathbf{s}_{\text{gt}}\mid O),
\end{equation}
with equality when $g(O)$ is $\mathbb{E}[\mathbf{s}_{\text{gt}}\mid O]$. Hence the Bayes optimal risk under $\mathcal{L}_{\text{Pearson}}$ is
\begin{equation}
R^{\star}(O)=\inf_{g}\big[1-\mathrm{Corr}(\mathbf{s}_{\text{gt}},g(O))\big] = 1-\eta(\mathbf{s}_{\text{gt}}\mid O).
\end{equation}
Assuming $f_{\theta}$ is expressive enough to attain $R^{\star}(O)$,
\begin{equation}
\begin{aligned}
R_{\text{pr}}^* &= 1-\eta(\mathbf{s}_{\text{gt}}\mid \mathbf{V},\mathbf{z}), \\
R_{\text{fr}}^* &= 1-\eta(\mathbf{s}_{\text{gt}}\mid \mathbf{V}),
\end{aligned}
\end{equation}
and thus
\begin{equation}
\label{supply_eq:R_eta}
R_{\text{fr}}^* - R_{\text{pr}}^* = \eta(\mathbf{s}_{\text{gt}}\mid \mathbf{V},\mathbf{z}) - \eta(\mathbf{s}_{\text{gt}}\mid \mathbf{V}).
\end{equation}

By the law of total variance,
\begin{equation}
\begin{aligned}
\eta^2(\mathbf{s}_{\text{gt}}\mid \mathbf{V},\mathbf{z}) - \eta^2(\mathbf{s}_{\text{gt}}\mid \mathbf{V})&= \frac{\mathbb{E}\big[\mathrm{Var}\big(\mathbb{E}[\mathbf{s}_{\text{gt}}\mid \mathbf{V},\mathbf{z}] \,\big|\, \mathbf{V}\big)\big]}{\mathrm{Var}(\mathbf{s}_{\text{gt}})} \\
& \ge 0,
\end{aligned}
\end{equation}
if
\begin{equation}
\label{supply_eq:P_E}
\mathbb{P}\,\big(\mathbb{E}[\mathbf{s}_{\text{gt}}\mid \mathbf{V},\mathbf{z}] \neq \mathbb{E}[\mathbf{s}_{\text{gt}}\mid \mathbf{V}]\big) > 0,
\end{equation}
then
\begin{equation}
\label{supply_eq:eta-eta}
\eta(\mathbf{s}_{\text{gt}}\mid \mathbf{V},\mathbf{z})-\eta(\mathbf{s}_{\text{gt}}\mid \mathbf{V}) > 0 ,
\end{equation}
substituting Eq. (\ref{supply_eq:eta-eta}) into Eq. (\ref{supply_eq:R_eta}),
\begin{equation}
\label{supply_eq:R*-R*}
R_{\text{fr}}^* - R_{\text{pr}}^* > 0,
\end{equation}
which implies that
\begin{equation}
\label{supply_eq:P>0-->R*-R*>0}
\mathbb{P}\,\big(\mathbb{E}[\mathbf{s}_{\text{gt}}\mid \mathbf{V},\mathbf{z}] \neq \mathbb{E}[\mathbf{s}_{\text{gt}}\mid \mathbf{V}]\big) > 0 \Longrightarrow R_{\text{fr}}^* - R_{\text{pr}}^* > 0.
\end{equation}

In particular, if
\begin{equation}
I(\mathbf{s}_{\text{gt}};\mathbf{z}\mid\mathbf{V})>0,
\end{equation}
then the conditional distributions $p(\mathbf{s}_{\text{gt}}\mid \mathbf{V},\mathbf{z})$ and $p(\mathbf{s}_{\text{gt}}\mid \mathbf{V})$ must differ on a set of positive probability, hence Eq. (\ref{supply_eq:P_E}) holds, i.e.,
\begin{equation}
\label{supply_eq:I>0-->P>0}
I(\mathbf{s}_{\text{gt}};\mathbf{z}\mid\mathbf{V})>0 \Longrightarrow \mathbb{P}\,\big(\mathbb{E}[\mathbf{s}_{\text{gt}}\mid \mathbf{V},\mathbf{z}] \neq \mathbb{E}[\mathbf{s}_{\text{gt}}\mid \mathbf{V}]\big) > 0,
\end{equation}
substituting Eq. (\ref{supply_eq:I>0-->P>0}) into Eq. (\ref{supply_eq:P>0-->R*-R*>0}),
\begin{equation}
I(\mathbf{s}_{\text{gt}};\mathbf{z}\mid\mathbf{V})>0 \Longrightarrow R_{\text{fr}}^* - R_{\text{pr}}^* > 0.
\end{equation}
Since this difference is strictly positive, there exists $\epsilon_{\text{info}}>0$ such that
\begin{equation}
R_{\text{fr}}^* - R_{\text{pr}}^* \ge \epsilon_{\text{info}}.
\end{equation}
\end{proof}

\subsection{Proof of Eq. (9)}
\label{appendix:proof of eq9}
Let $\mathfrak{R}_T(\mathcal{F}_{\text{pr}})$ and $\mathfrak{R}_T(\mathcal{G})$ denote the privileged predictor's and the consistency function's complexities. Then the following generalization bound hold with probability at least $1-\gamma$:
\begin{equation}
\begin{aligned}
R_{\text{fr}}(\theta)\le &\;\hat R_{\text{pr}}(\theta)+ 4L\,\mathfrak{R}_T(\mathcal{F}_{\text{pr}})+ 3\sqrt{\tfrac{\log(2/\gamma)}{2T}} \\
&\;+ L\,\sqrt{\hat C(\theta)+ 2\mathfrak{R}_T(\mathcal{G})+ 3\sqrt{\tfrac{\log(2/\gamma)}{2T}}}.  
\end{aligned}
\end{equation}

\begin{proof}
By the Cauchy–Schwarz inequality,
\begin{equation}
\delta(\theta) = \mathbb{E}_\mathbf{V}[\Delta_\theta(\mathbf{V})] \le \sqrt{\mathbb{E}_\mathbf{V}[\Delta_\theta(\mathbf{V})^2]}.
\end{equation}
Let 
\begin{equation}
\mathcal{G} =\bigl\{g_\theta(\mathbf{V}) = \|f_\theta(\mathbf{V}, \mathbf{z}) - f_\theta(\mathbf{V}, \mathbf{0})\|_2^2:\theta \in \Theta\bigr\}.
\end{equation}
Each $g_\theta$ maps an input $\mathbf{V}$ to its squared consistency error. Its expectation is
\begin{equation}
\mathbb{E}_\mathbf{V}[g_\theta(\mathbf{V})] = \mathbb{E}_\mathbf{V}\,[\|f_\theta(\mathbf{V}, \mathbf{z}) - f_\theta(\mathbf{V}, \mathbf{0})\|_2^2].
\end{equation}
For any fixed sample set \(S = \{\mathbf{V}^1,\dots,\mathbf{V}^T\}\), the empirical Rademacher complexity \cite{bartlett2002rademacher} of $\mathcal{G}$ is defined as
\begin{equation}
\mathfrak{R}_S(\mathcal{G})=\mathbb{E}_{\sigma}\left[\sup_{g \in \mathcal{G}}\frac{1}{T}\sum_{t=1}^T\sigma_t g(\mathbf{V}^t)\right],
\end{equation}
where $\sigma_i \in \{-1,+1\}$ are i.i.d. Rademacher variables. Taking expectation over all possible samples $S$ drawn from the underlying distribution yields the expected Rademacher complexity,
\begin{equation}
\mathfrak{R}_T(\mathcal{G})=\mathbb{E}_S\big[\mathfrak{R}_S(\mathcal{G})\big].
\end{equation}
A standard uniform convergence theorem based on Rademacher complexity states that for any $\gamma \in (0,1)$, with probability at least $1 - \gamma$ over the draw of $S$,
\begin{equation}
\begin{aligned}
\label{supply_eq:Rademacher complexity}
&\forall g \in \mathcal{G}, \\
&\mathbb{E}_\mathbf{V}[g(\mathbf{V})]\le\frac{1}{T}\sum_{t=1}^T g(\mathbf{V}^t)+ 2\,\mathfrak{R}_T(\mathcal{G})+ 3\sqrt{\frac{\log(2/\gamma)}{2T}}.
\end{aligned}
\end{equation}
Substituting $g_\theta(\mathbf{V}) = \|f_\theta(\mathbf{V}, \mathbf{z}) - f_\theta(\mathbf{V}, \mathbf{0})\|_2^2 = \Delta_\theta(\mathbf{V})^2$ and $\frac{1}{T}\sum_t g_\theta(\mathbf{V}^t) = \hat C(\theta)$ into Eq. (\ref{supply_eq:Rademacher complexity}) gives
\begin{equation}
\mathbb{E}_\mathbf{V}[\Delta_\theta(\mathbf{V})^2] \le \hat C(\theta) + 2\mathfrak{R}_T(\mathcal{G}) + 3\sqrt{\frac{\log(2/\gamma)}{2T}}.
\end{equation}
Thus, we obtain
\begin{equation}
\label{supply_eq:delta_theta}
\delta(\theta)\le\sqrt{\hat C(\theta) + 2\,\mathfrak{R}_T(\mathcal{G}) + 3\sqrt{\frac{\log(2/\gamma)}{2T}}}.
\end{equation}

Let \(\mathcal{F}_{\text{pr}} = \bigl\{ (\mathbf{V},\mathbf{z}) \mapsto f_\theta(\mathbf{V},\mathbf{z}) : \theta \in \Theta \bigr\}\) denote the class of all possible mappings realized by the model $f_\theta$ when both the visual input $\mathbf{V}$ and the privileged information $\mathbf{z}$ are provided. Correspondingly, we define the induced loss function class \(\mathcal{L}_{\text{pr}}=\bigl\{(x,z,y) \mapsto \ell(f_\theta(x,z), y) : \theta \in \Theta\bigr\},\) where $\ell(\cdot,y)$ is $L$-Lipschitz in its first argument. By vector-contraction inequality \cite{maurer2016vector},
\begin{equation}
\mathfrak{R}_T(\mathcal{L}_{\text{pr}}) \le 2L \cdot \mathfrak{R}_T(\mathcal{F}_{\text{pr}}).
\end{equation}
By Eq. (\ref{supply_eq:Rademacher complexity}), we have
\begin{equation}
R_{\text{pr}}(\theta)\le\hat R_{\text{pr}}(\theta)+ 2\mathfrak{R}_T(\mathcal{L}_{\text{pr}})+ 3\sqrt{\frac{\log(2/\gamma)}{2T}}.
\end{equation}
Therefore, with probability at least $1-\gamma$:
\begin{equation}
\label{supply_eq:R_pr_theta}
R_{\text{pr}}(\theta)\le \hat R_{\text{pr}}(\theta)+ 4L \cdot \mathfrak{R}_T(\mathcal{F}_{\text{pr}})+ 3\sqrt{\frac{\log(2/\gamma)}{2T}}.
\end{equation}

Substituting Eq. (\ref{supply_eq:delta_theta}) and Eq. (\ref{supply_eq:R_pr_theta}) into Eq. (\ref{supply_eq:Lipschitz}) yields
\begin{equation}
\begin{aligned}
\label{supply_eq:fr_pr}
R_{\text{fr}}(\theta)\le &\;\hat R_{\text{pr}}(\theta)+ 4L\,\mathfrak{R}_T(\mathcal{F}_{\text{pr}})+ 3\sqrt{\tfrac{\log(2/\gamma)}{2T}} \\
&\;+ L\,\sqrt{\hat C(\theta)+ 2\mathfrak{R}_T(\mathcal{G})+ 3\sqrt{\tfrac{\log(2/\gamma)}{2T}}}.  
\end{aligned}
\end{equation}
\end{proof}

\end{document}